\pdfoutput=1
\documentclass[11pt]{article}

\usepackage[preprint]{acl}

\usepackage{times}
\usepackage{latexsym}
\usepackage[T1]{fontenc}
\usepackage[utf8]{inputenc}
\usepackage{microtype}
\usepackage{inconsolata}
\usepackage{graphicx}
\usepackage{amsmath}
\usepackage{amssymb}
\usepackage{booktabs}
\usepackage{multirow}
\usepackage{tikz}
\usetikzlibrary{arrows.meta,positioning,calc,fit}
\usepackage{pgfplots}
\pgfplotsset{compat=1.17}
\usepackage{listings}
\usepackage{tcolorbox}
\tcbuselibrary{listings,breakable}
\definecolor{promptfill}{HTML}{F2F5F8}
\definecolor{promptframe}{HTML}{8F9BA8}
\definecolor{prompttitle}{HTML}{E9EEF3}
\definecolor{promptink}{HTML}{263647}
\newtcblisting{promptbox}[1]{listing only,breakable,colback=promptfill,colframe=promptframe,colbacktitle=prompttitle,coltitle=promptink,fonttitle=\footnotesize\bfseries,title={#1},boxrule=0.4pt,arc=1pt,left=3pt,right=3pt,top=1pt,bottom=1pt,toptitle=1pt,bottomtitle=1pt,before skip=4pt,after skip=6pt,listing options={basicstyle=\footnotesize\ttfamily,breaklines=true,columns=fullflexible,keepspaces=true}}

\makeatletter
\newcommand{\fullcolumn}{\global\@colroom\@colht\global\vsize\@colht}
\makeatother

\title{When Reasoning Traces Conflict with Evidence: \\How Answer Configuration Shapes Evidence Following}

\author{Shaoang Li\thanks{\ Equal contribution.} \quad Yanhang Shi\footnotemark[1] \quad Jian Li \\
  Stony Brook University \\
  \texttt{\{shaoang.li, yanhang.shi, jian.li.3\}@stonybrook.edu}}

\begin{document}
\maketitle

\begin{abstract}
A reasoning trace retained at answer time plays two roles: it can raise accuracy, and it can compete with the current evidence when the two disagree. We disentangle these roles by crossing verified evidence versions with the traces generated from each and answering from the same traces and evidence under two configurations: continuation of the model's own thinking and a separate answer call. Across three model families (gpt-oss, Gemma4, Qwen3) on HotpotQA and MuSiQue, the separate call raises evidence-following over continuation by 29.9 to 60.6 points. On the same pairs, removing the trace lowers accuracy by similar amounts under both configurations. Yet continuation repeats the stale conclusion in most conflicts, and the separate call follows the current evidence far more often. Benefiting from a trace and deferring to it are therefore distinct properties. Further analyses show that the final line of a trace carries most of its value, gold evidence helps most before the trace is written, distractors weaken correction under a separate call, and renewed reasoning, unlike a verification instruction, follows the current evidence in most conflicts. Because trace deletion alone cannot reveal which source an answer follows, evaluations should report both properties together with the answer configuration.
\end{abstract}

\section{Introduction}
\label{sec:intro}

Large language models can generate explicit reasoning traces before committing to a final answer \citep{wei2022cot}. Retained at answer time, such a trace can inform later answer generation, carrying forward selected facts and derived conclusions of the reasoning stage \citep{tac2026,traceasstate2026}. In multi-stage workflows, however, the answer-time evidence may differ from the evidence the trace was generated from, so the trace can preserve a conclusion that the current evidence contradicts. Systems that reuse earlier reasoning must therefore retain the useful information in a trace while remaining responsive to updated evidence.

We characterize the answer stage by two measurable properties. \emph{Trace use} is the effect of the supplied trace on later answers: removing, shortening, or replacing it can change the answer and its accuracy \citep{lanham2023measuring,paul2024frodo,case2026}. \emph{Conflict response} is which source the answer follows when the trace and the current evidence disagree. The former quantifies how earlier reasoning contributes to answer generation, the latter how the answer stage arbitrates between conflicting sources. Our central question is:
\begin{quote}
\itshape Does an answer that uses its trace also defer to it when current evidence disagrees, and how does the answer configuration shape this choice?
\end{quote}

\begin{figure*}[t]
\centering
\includegraphics{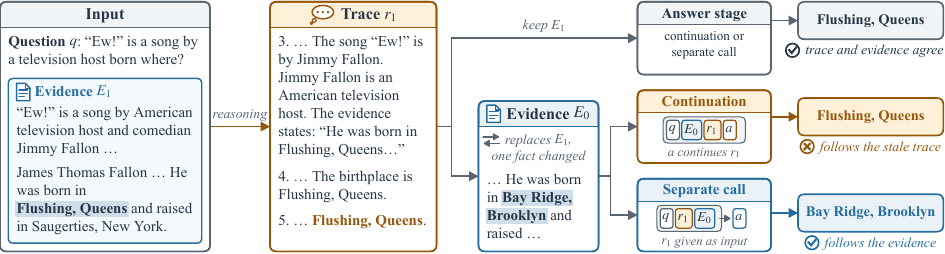}
\caption{A crossed conflict on a HotpotQA question, with the outputs of Gemma4-E4B (evidence and trace abridged). $E_0$ is the original evidence, $E_1$ an edited version that changes the birthplace of Jimmy Fallon, and the trace $r_1$ is generated from $E_1$. When $E_0$ replaces $E_1$ at answer time, continuation repeats the stale conclusion of $r_1$, while a separate call on the identical trace follows the current evidence $E_0$.}
\label{fig:example}
\end{figure*}

Figure~\ref{fig:example} illustrates the conflict-response half of this question for two answer configurations: \emph{continuation} of the model's own thinking from a stored trace, and a \emph{separate answer call} that receives the trace as input. Gemma4-E4B first generates a trace from an edited evidence passage. With the original evidence supplied at answer time, continuation repeats the stale conclusion of the trace, but the separate call follows the current evidence. Because the model, question, trace text, and evidence content are held fixed, the divergence is attributable to the answer configuration.

Prior work has approached these properties along separate lines: trace interventions measure how reasoning traces influence model outputs \citep{lanham2023measuring,paul2024frodo,zaman2025causallens}, and work on changed evidence lets the model reason again after the change \citep{shi2026visualswap}. We connect the two at the answer stage, where a previously generated trace and the current evidence can support competing answers, and measure both properties on identical traces and evidence. In this setting, the retained trace acts as a third source in knowledge conflicts, alongside retrieved context and parametric memory \citep{longpre2021entity,xie2024adaptive}.

Our experiments show that the answer configuration has a large effect on conflict response. Across gpt-oss-20b, Gemma4-E4B, and Qwen3-14B on HotpotQA \citep{hotpotqa} and MuSiQue \citep{musique}, a separate answer call raises evidence-following over continuation by 29.9 to 60.6 points on identical traces and evidence pairs, with every pair-level 95\% bootstrap interval well above zero. Removing the trace from the same pairs lowers accuracy under both configurations, and with gold evidence the two losses differ by at most 4.7 points. Benefiting from a reasoning trace and deferring to its conclusion when the evidence disagrees are therefore distinct properties.

Our contributions are:
\begin{itemize}
\item \textbf{A joint evaluation of trace use and conflict response.} We combine trace interventions with crossed conflicts between verified evidence versions and the traces generated from them. On the same pairs, we measure both how much a trace contributes to accuracy and which source an answer follows when trace and evidence disagree.

\item \textbf{A large answer-configuration effect that trace deletion cannot detect.} For three model families, a separate answer call follows the current evidence far more often than continuation, although removing the trace costs both configurations comparable accuracy. Input order accounts for most of this gap in two of the three models.

\item \textbf{Evidence on what traces carry and on the limits of correction.} The final line of a trace retains most of its value, gold evidence helps most before the trace is written, and distractors weaken correction under a separate call. Renewed reasoning corrects most conflicts, but an instruction to verify the trace does not.
\end{itemize}

\section{Experimental Framework}
\label{sec:framework}

Our protocol decouples trace generation from answer generation, so that the trace and the answer-time evidence can be varied independently. Removing or editing a stored trace measures how much the answer draws on it, and pairing the evidence with a trace generated from the other evidence version identifies which source the answer follows. In every conflict comparison, both answer configurations receive identical stored traces and evidence.

\subsection{Stages and Answer Configurations}
\label{sec:inference}

Each question $q$ has a context $C=E\cup D$, where $E$ comprises the annotated supporting paragraphs and $D$ the distractor paragraphs. We call $C$, $E$, and $D$ Full, Gold, and Distr., respectively. A reasoning stage generates a trace $r\sim p_\theta(r\mid q,C_1)$, and an answer stage then generates $a\sim p_\theta(a\mid q,C_2,\tilde r)$, where $\tilde r$ is the version of the trace supplied at answer time. Interventions draw $C_1$ and $C_2$ from Full, Gold, Distr., and the empty context, or from the two evidence versions of the crossed conflicts, and retain, shorten, or remove $r$.

\paragraph{Continuation.}
Continuation reconstructs the reasoning prompt with the question and the selected answer-time evidence, inserts the stored trace into the model's own thinking block, closes that block, and generates a short final answer. The evidence thus precedes the trace, and the answer is generated immediately after the supplied thinking text.

\paragraph{Separate answer call.}
The separate call sends a new request with thinking off. Its user message asks for a short answer and supplies the question, the trace, and then the evidence, so the trace enters as user-supplied text and the evidence comes last. The configurations thus differ in three respects: where the trace sits (in the thinking block, or in the user message with thinking off), which input comes last, and which instruction is given. Section~\ref{sec:components} separates input order from the other two.

Standard inference uses $C_1=C_2=C$ and $\tilde r=r$. We write $X\rightarrow Y$ when the trace is generated from context $X$ and the answer from context $Y$. Each trace, including the gold-written and distractor-written traces of the cross-stage conditions, is generated once and reused under every answer-time context.

\begin{table}[t]
\centering
\scriptsize
\setlength{\tabcolsep}{3pt}
\begin{tabular}{llll}
\toprule
Model & Data & Ablations & Conflicts \\
\midrule
Qwen3-4B, 8B, 32B & H / M & Cont. & Cont. \\
Qwen3-14B & H / M & Cont. & Cont. / Sep. \\
Qwen3.5-4B & H / M & Sep. & none \\
Gemma4-E4B & H / M & Sep. / Cont.$^{a}$ & Cont. / Sep. \\
gpt-oss-20b & H / M & Cont. / Sep. & Cont. / Sep. \\
DeepSeek-V4-Pro & H & Sep.$^{b}$ & Sep. \\
\bottomrule
\end{tabular}
\caption{Answer configurations tested for each model. H is HotpotQA and M is MuSiQue. Cont.\ is continuation and Sep.\ a separate answer call. Ablations comprise the answer-time, reasoning-time, and cross-stage conditions. $^{a}$The continuation runs reuse the separate-call traces and are reported only as paired contrasts (Appendix~\ref{app:full}). $^{b}$Only the answer-time ablations.}
\label{tab:coverage}
\end{table}

\subsection{Crossed Trace-Evidence Conflicts}
\label{sec:interventions}

We construct two versions of the supporting evidence, the original $E_0$ and an edited $E_1$, with target answers $a_0\neq a_1$. Following entity-substitution studies of knowledge conflicts \citep{longpre2021entity}, we obtain $E_1$ by replacing the gold answer $a_0$ throughout the gold paragraphs with a proposed counterfactual answer $a_1$ of the same type, and keep a pair only if a verifier model recovers the target answer of each version. Verification yields 683 pairs from HotpotQA bridge questions and 417 from MuSiQue two-hop questions.

For each model, we generate $r_0$ from $E_0$ and $r_1$ from $E_1$ and retain pairs in which each trace states the target of its own evidence version and not the other target. We then evaluate all four pairings $(E_i,r_j)$, with $i,j\in\{0,1\}$, once with gold evidence alone and once with the original distractors added. Matched cells $(E_i,r_i)$ measure accuracy when both sources support $a_i$, and crossed cells $(E_i,r_j)$, $i\neq j$, reveal whether the answer follows the evidence ($a_i$) or the trace ($a_j$). Three models, gpt-oss-20b, Gemma4-E4B, and Qwen3-14B, answer under both configurations, and every other model except Qwen3.5-4B answers under one (Table~\ref{tab:coverage}).

Here, \emph{correction} means agreement with the verified target of the current evidence. Because each model retains a different set of pairs, comparisons across models also reflect these sets.

\begin{figure*}[t!]
\centering
\includegraphics{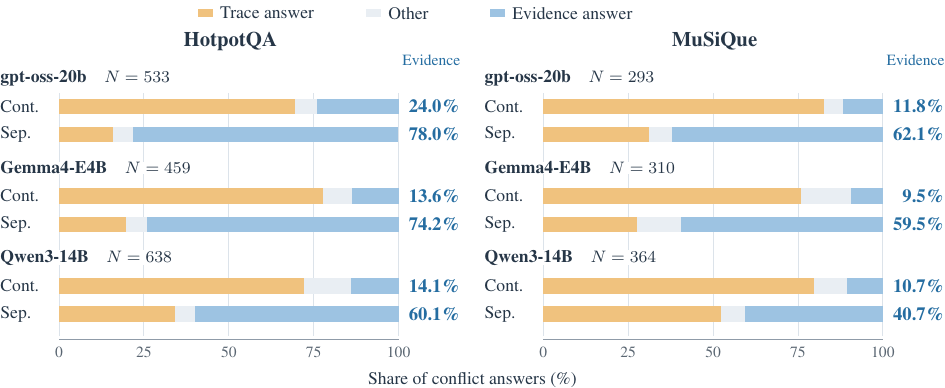}
\caption{The two answer configurations on identical crossed-conflict pairs, traces, and gold evidence. Each bar aggregates the two off-diagonal cells $(E_i,r_j)$, $i\ne j$, and the right-hand labels give evidence-following. $N$ is the number of evidence pairs retained for that model. Appendix~\ref{app:full} adds matched-cell accuracy, the other Qwen3 models, and paired bootstrap intervals for the gain of the separate call.}
\label{fig:conflict}
\end{figure*}

\subsection{Evidence and Trace Interventions}
\label{sec:diagnostic_design}

\paragraph{Answer-time ablations.}
Given a trace generated from Full, the answer stage receives the full trace, approximately half of its steps chosen at random, or the no-trace placeholder, with Full, Gold, Distr., or empty answer-time evidence. Additional controls retain only the final trace line, remove that line, substitute a length-matched gold extractive summary, supply an unrelated trace, or leave the trace field empty.

\paragraph{Reasoning-time ablations.}
Both stages receive the same reduced context, $C_1=C_2\in\{E,D,\emptyset\}$. Comparing Gold$\rightarrow$Gold with Full$\rightarrow$Gold holds the answer-time evidence fixed and varies only the evidence from which the trace is generated, thereby isolating the value of removing distractors before the trace is written. Without evidence, trace and answer come from a single call.

\paragraph{Cross-stage substitution and repair.}
Distr.$\rightarrow$Gold supplies annotated evidence only after a trace has been generated from distractors, and Gold$\rightarrow$Distr.\ tests how well a useful trace retains its value once the supporting paragraphs are removed at answer time. We distinguish repair of standard Full$\rightarrow$Full failures from repair of Distr.$\rightarrow$Distr.\ failures. Rewriting replaces the trace with one generated from Gold, without showing the model the earlier trace, so a rewritten answer is the Gold$\rightarrow$Gold answer on the baseline failures.

\subsection{Datasets, Models, and Coverage}
\label{sec:setup}

We use the distractor setting of HotpotQA, which embeds two supporting paragraphs among ten, and the answerable portion of MuSiQue, which embeds two to four among twenty \citep{hotpotqa,musique}. The main ablations use the first 2,000 validation questions of each dataset. MuSiQue contexts are cut at 12,000 characters, which removes at least one supporting paragraph from the Full context of 171 questions (8.6\%). Excluding these questions leaves the conclusion of Section~\ref{sec:enters} unchanged (Appendix~\ref{app:setup}).

The study spans eight models from four families: Qwen3-4B, Qwen3-8B, Qwen3-14B, Qwen3-32B \citep{qwen3}, Qwen3.5-4B \citep{qwen35}, instruction-tuned Gemma4-E4B \citep{gemma4}, gpt-oss-20b \citep{gptoss2025}, and DeepSeek-V4-Pro \citep{deepseekv4}, a larger model that we query through its API on smaller HotpotQA question sets (Table~\ref{tab:coverage}).

Qwen3.5-4B and Qwen3-32B return empty traces for 61.2\% and 38.9\% of the MuSiQue questions, and the other Qwen3 sizes and Gemma4-E4B for under 4\%. Appendix~\ref{app:setup} gives the decoding settings, length limits, and the question sets of analyses that need a trace, and Appendix~\ref{app:prompts} the prompts.

\subsection{Metrics and Comparison Units}
\label{sec:evaluation}

We report exact match (EM), token F1, and \emph{Contains}, which accepts an answer when the target occurs in it as a contiguous sequence of whole words. \emph{Trace contribution} is the accuracy with the trace minus the accuracy with the no-trace placeholder on the same questions and answer-time evidence. It measures the net effect of the trace, since removing a trace can turn correct answers into errors and vice versa. The \emph{answer change rate} is the share of normalized outputs that differ from the corresponding standard answer, and a \emph{repair rate} is the share of a baseline failure set that becomes correct under an intervention.

In crossed conflicts, an answer is \emph{evidence-following} when it matches $a_i$ alone and \emph{trace-following} when it matches $a_j$ alone. Other covers answers that match both targets or neither. Matching uses EM or Contains, and MuSiQue aliases are accepted for the original answer, which changes these rates by at most 1.8 points. Matched-cell accuracy is scored against the shared target. Rates average the two conflict directions.

Intervals are 95\% bootstrap intervals over questions in the ablations and evidence pairs in the crossed conflicts. Ablation contrasts also report exact McNemar tests.

\section{Trace Use and Conflict Response Dissociate}
\label{sec:answerstep}

With the model, trace, and evidence held fixed, the answer configuration strongly changes which source an answer follows in a conflict, although both configurations gain comparable accuracy from the trace (Sections~\ref{sec:separate} and~\ref{sec:two_dimensions}). Continuation repeats the stale conclusion in most conflicts for every tested model, and distractors weaken the correction that a separate call enables (Sections~\ref{sec:continuation} and~\ref{sec:noise_boundary}).

\subsection{A Separate Call Follows Current Evidence}
\label{sec:separate}

On identical conflicts, a separate answer call follows the current evidence far more often than continuation (Figure~\ref{fig:conflict}). For gpt-oss-20b on HotpotQA, evidence-following rises from 24.0\% under continuation to 78.0\% under a separate call. Across gpt-oss-20b and Gemma4-E4B on both datasets, the gain is 50.0 to 60.6 points, and every pair-level 95\% bootstrap interval lies far above zero. The gain holds in both conflict directions. Even when following the evidence means giving the edited answer, which may contradict what the model knows, the gain for these two models is 45.1 to 59.0 points. The effect extends to a third model family: with a separate call on its own retained pairs and traces, the evidence-following of Qwen3-14B rises from 14.1\% to 60.1\% on HotpotQA and from 10.7\% to 40.7\% on MuSiQue.

This correction costs little when trace and evidence agree. For gpt-oss-20b and Gemma4-E4B, both configurations answer 87.7\% to 96.7\% of the matched cells correctly, with the separate call at most 1.9 points lower. Qwen3-14B is even more accurate under the separate call.

\subsection{Trace Contribution Is Comparable}
\label{sec:two_dimensions}

Unlike conflict response, trace use is comparable across the two configurations. To compare both measurements on the same inputs, we also answer the retained pairs of gpt-oss-20b, Gemma4-E4B, and Qwen3-14B without their traces. With gold evidence, the trace raises matched-cell accuracy under both configurations, and the two contributions differ by at most 4.7 points for each model and dataset, but evidence-following differs by 29.9 to 60.6 points (Table~\ref{tab:trace_use_conflict}). On the original questions with full contexts, the trace also raises the EM of gpt-oss-20b under both configurations, by 7.7 points under continuation and 10.8 under a separate call on HotpotQA, and by 17.0 and 18.4 on MuSiQue.

\begin{table}[t]
\centering
\footnotesize
\setlength{\tabcolsep}{4pt}
\begin{tabular}{lcccc}
\toprule
& \multicolumn{2}{c}{Trace contribution} & \multicolumn{2}{c}{Evid. (\%)} \\
\cmidrule(lr){2-3}\cmidrule(lr){4-5}
Model & Cont. & Sep. & Cont. & Sep. \\
\midrule
\multicolumn{5}{l}{\textit{HotpotQA}} \\
gpt-oss-20b & 4.7 & 4.7 & 24.0 & 78.0 \\
Gemma4-E4B & 5.2 & 8.3 & 13.6 & 74.2 \\
Qwen3-14B & 6.6 & 3.7 & 14.1 & 60.1 \\
\midrule
\multicolumn{5}{l}{\textit{MuSiQue}} \\
gpt-oss-20b & 13.7 & 11.1 & 11.8 & 62.1 \\
Gemma4-E4B & 14.4 & 19.0 & 9.5 & 59.5 \\
Qwen3-14B & 14.1 & 9.6 & 10.7 & 40.7 \\
\bottomrule
\end{tabular}
\caption{Trace use and conflict response on the same retained pairs and traces, with gold evidence only. Trace contribution is the matched-cell accuracy with the trace minus that without it, in points. Evid.\ is evidence-following in the crossed cells. Table~\ref{tab:joint} adds pair counts and the distractor setting.}
\label{tab:trace_use_conflict}
\end{table}

The two measurements therefore dissociate: a trace-deletion test would find similar trace use under the two configurations, and only the conflict test reveals that they resolve conflicts in markedly different ways. How much an answer gains from a trace says little about whether it defers to that trace when the evidence disagrees. With the original distractors added, the trace contributes more under both configurations (11.1 to 22.5 points on HotpotQA and 34.5 to 45.5 on MuSiQue), and the same dissociation holds.

\subsection{Continuation Retains the Stale Conclusion}
\label{sec:continuation}

For every tested model, continuation repeats the conclusion of the stale trace in 65.6\% to 87.0\% of crossed answers. Model scale attenuates this only modestly: on the pairs that all four Qwen3 sizes retain, evidence-following rises from 8.7\% at 4B to 21.4\% at 32B on HotpotQA and from 4.3\% to 18.6\% on MuSiQue, gains of 12.7 and 14.3 points, less than half of the 46.0 and 29.9 points that a separate call adds for Qwen3-14B on its own retained pairs.

\subsection{Evidence Quality Limits Correction Under Separate Calls}
\label{sec:noise_boundary}

Distractors weaken the correction that a separate call provides. On HotpotQA, gpt-oss-20b and Gemma4-E4B still follow the evidence in 71.0\% and 73.9\% of conflicts with distractors added, but for Qwen3-14B, evidence-following falls from 60.1\% to 45.5\%. On MuSiQue, it falls from 62.1\% to 41.3\% for gpt-oss-20b, from 40.7\% to 25.3\% for Qwen3-14B, and from 59.5\% to 52.7\% for Gemma4-E4B. With distractors, Qwen3-14B on both datasets and gpt-oss-20b on MuSiQue follow the stale trace more often than the evidence, and Qwen3-14B already does so on MuSiQue with gold evidence alone.

DeepSeek-V4-Pro, which we query with a separate answer call on HotpotQA, also relies on its trace. Removing the trace changes 23.9\% of its answers, and removing all evidence changes only 6.9\%. It follows the evidence in 58.4\% of conflicts with gold evidence and in 53.4\% with distractors added (Appendix~\ref{app:deepseek}).

The answer configuration thus largely determines whether the current evidence can override a stale trace, and the quality of that evidence bounds how often it does.

\begin{figure*}[t!]
\centering
\includegraphics{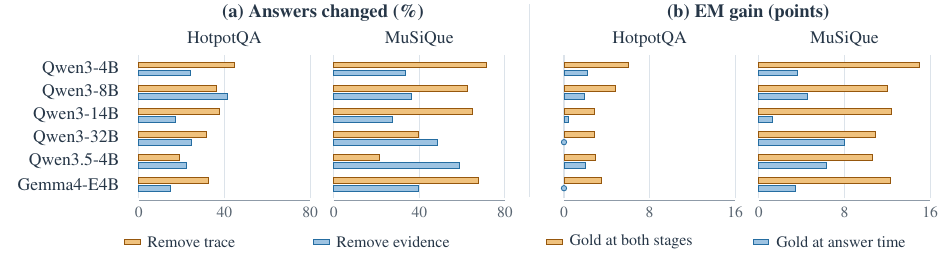}
\caption{Stage-separated ablations. (a) Share of answers that change after removing the trace or all evidence at answer time, relative to the full-input answer. (b) EM gains over Full$\rightarrow$Full when gold evidence is used at both stages (Gold$\rightarrow$Gold) or only at answer time (Full$\rightarrow$Gold). Appendix~\ref{app:full} gives the underlying tables.}
\label{fig:ablations}
\end{figure*}

\section{Trace Content, Evidence Quality, and Repair}
\label{sec:trace}

On the original questions, the answer stage draws mainly on the conclusion of a trace, and a wrong conclusion written into the trace often overrides the evidence (Section~\ref{sec:follow}). Evidence helps most when it arrives before that conclusion is written (Section~\ref{sec:enters}), and once a trace exists, replacing it with a trace written from gold evidence repairs more failures than supplying better evidence beside it (Section~\ref{sec:repair}).

\subsection{The Final Line Retains Most of the Value}
\label{sec:follow}

Removing the trace lowers EM for all six models of Figure~\ref{fig:ablations}, and in eight of the twelve model-dataset combinations it changes more answers than removing all evidence, for example 44.8\% against 24.0\% for Qwen3-4B on HotpotQA (Figure~\ref{fig:ablations}a). The four exceptions are Qwen3-8B on HotpotQA and three settings with many empty traces: Qwen3.5-4B on both datasets and Qwen3-32B on MuSiQue.

The final line of the trace carries most of this value. For Qwen3-4B and Qwen3-14B, the final line alone stays within 2.0 EM points of the full trace on both datasets, while removing it costs 2.9 to 8.5 points (Appendix~\ref{app:audits}). The answer stage thus draws mainly on the conclusion of earlier reasoning, precisely what a stale trace gets wrong.

A short text stating the required facts can substitute for the trace: a length-matched summary of the gold supporting sentences, selected here with the gold annotations, stays within 1.5 EM points of the trace on HotpotQA and exceeds it by 2.2 to 2.7 points on MuSiQue. Conversely, the answer stage copies wrong content from the trace. When the gold answer in a trace is replaced by the title of a distractor paragraph, 39.9\% to 63.6\% of previously correct answers become wrong across the four Qwen3 sizes, and in 27.5\% to 41.8\% of the edited cases the model answers with the inserted title, even with the full evidence present. As the Qwen3 models answer by continuation, this edit reproduces the stale-trace conflict of Section~\ref{sec:answerstep} on the original questions.

\subsection{Reasoning-Time Evidence Helps Most}
\label{sec:enters}

\begin{figure*}[t]
\centering
\includegraphics{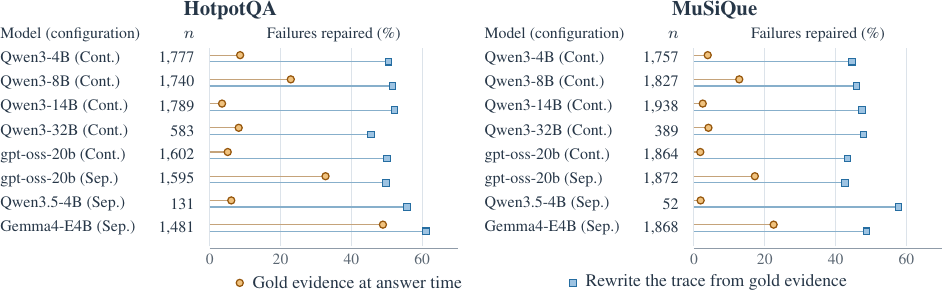}
\caption{Repair of distractor-written failures. Circles give the repair rate when gold evidence replaces the distractors at answer time, and squares the higher rate after rewriting the trace from gold evidence. Stems start at zero, and $n$ is the number of baseline failures per row.}
\label{fig:repair}
\end{figure*}

Gold$\rightarrow$Gold improves EM over Full$\rightarrow$Full for all six models, and on both datasets the gain exceeds that from removing distractors only at answer time (Figure~\ref{fig:ablations}b). For Qwen3-8B on MuSiQue, EM is 35.0 for Full$\rightarrow$Full, 39.5 for Full$\rightarrow$Gold, and 47.0 for Gold$\rightarrow$Gold. With the answer-time gold evidence held fixed, generating a new trace from that evidence yields the larger gain.

Traces written from distractors alone rarely reach the gold answer: for the six models, at most 23.0\% of them contain it, and the answers based on them score at most 15.0 EM on HotpotQA.

\paragraph{Repairing full-context failures.}
Among questions answered incorrectly under Full$\rightarrow$Full, regenerating the trace from Gold repairs more failures than removing the distractors at answer time in all sixteen combinations of model, dataset, and answer configuration, for example 18.3\% against 8.9\% for Qwen3-4B on HotpotQA.

\subsection{Rewriting Is the Stronger Repair}
\label{sec:repair}

Rewriting the trace from Gold repairs 42.6\% to 61.0\% of the Distr.$\rightarrow$Distr.\ failures for every model and configuration, while supplying Gold beside the distractor-written trace repairs at most 22.9\% under continuation and up to 48.9\% under a separate call (Figure~\ref{fig:repair}). Because both gpt-oss-20b configurations answer from the same traces, we also compare them on the questions where both fail. On these HotpotQA questions, rewriting repairs 49.8\% under continuation and 49.7\% under a separate call, while gold evidence at answer time repairs 4.1\% and 32.4\%. MuSiQue shows the same pattern. The configuration thus changes repair only while the distractor-written trace is kept.

\paragraph{Keeping a useful trace.}
A separate call pays little for its openness to the evidence when the trace is right and the evidence is poor. Compared with Gold$\rightarrow$Gold, answering from a gold-written trace with only the distractors (Gold$\rightarrow$Distr.) lowers the EM of gpt-oss-20b by 5.1 points under continuation and 7.4 under a separate call on HotpotQA, and by 3.2 and 3.3 on MuSiQue.

\section{Further Analyses and Implications}
\label{sec:implications}

Two further analyses ask how a stale trace can be corrected and what drives the gap between the two configurations. For Qwen3-14B, renewed reasoning follows the current evidence in most conflicts, but an instruction to verify the trace does not (Section~\ref{sec:whichpart}). For two of the three models, input order accounts for most of the gap (Section~\ref{sec:components}). Section~\ref{sec:practice} turns to the implications for evaluation and practice.

\subsection{Verification Instruction Versus Renewed Reasoning}
\label{sec:whichpart}

We test two further corrections of a stale trace with Qwen3-14B on its 638 retained HotpotQA pairs. \emph{Verify} keeps the trace in the thinking block and adds an instruction to check the trace against the evidence and to follow the evidence if they disagree. \emph{Re-reason} starts a new reasoning call over the current evidence and presents the stale trace as previous notes that may be outdated or wrong.

\begin{table}[t]
\centering
\footnotesize
\setlength{\tabcolsep}{4pt}
\begin{tabular}{lrrrr}
\toprule
Method & Trace & Evidence & Other & Tokens \\
\midrule
Continuation & 72.2 & 14.1 & 13.7 & 76.8 \\
Verify & 73.6 & 12.9 & 13.5 & 77.0 \\
Re-reason & 4.4 & 86.6 & 9.0 & 145.4 \\
\bottomrule
\end{tabular}
\caption{Correction procedures for Qwen3-14B on HotpotQA with gold evidence only. Percentages describe the two crossed outputs of each of the 638 evidence pairs, with the matching rule of Section~\ref{sec:evaluation}. Tokens is the mean number of generated tokens per crossed answer, including the reused trace of 68.1 tokens on average (Qwen3-14B tokenizer).}
\label{tab:steps}
\end{table}

The verification instruction we tested does not loosen the hold of the trace: Verify leaves trace-following at 73.6\%, close to the 72.2\% of continuation (Table~\ref{tab:steps}). A separate answer call follows the evidence in 60.1\% of the same conflicts, and renewed reasoning does better still: Re-reason follows the evidence in 86.6\% of conflicts, close to the 90.2\% accuracy of the original reasoning calls on the same evidence. Re-reason roughly doubles the generated text, from 76.8 to 145.4 tokens per crossed answer, because it adds a new trace and answer to the reused trace.

\subsection{What Drives the Configuration Effect}
\label{sec:components}

The configurations differ in trace placement, input order, and instruction. Variants of the separate call that change only the input order or the label of the trace show that input order accounts for most of the gap in two of the three models (Appendix~\ref{app:components}). When the separate call lists the evidence before the trace, so that the trace comes last as in continuation, evidence-following falls by 36.4 and 31.3 points for Gemma4-E4B and Qwen3-14B on HotpotQA and by 36.6 and 27.2 on MuSiQue. This single change closes 60\% to 91\% of their gap to continuation across both datasets. For gpt-oss-20b it costs only 7.2 points on HotpotQA and 20.3 on MuSiQue, so most of its gap comes from the trace placement and the reasoning instruction, which change together. Labeling the trace as the model's own earlier reasoning does not raise reliance on it. Nor does answer length explain the gap: both configurations give answers with a median of 4 to 5 tokens in the crossed conflicts.

\subsection{Implications for Evaluation and Practice}
\label{sec:practice}

Because trace-deletion results do not reveal which source an answer follows in a conflict (Section~\ref{sec:two_dimensions}), evaluations of evidence-based answers should report trace contribution, conflict response, and matched-cell accuracy together with the answer configuration. Reporting the configuration means stating the trace position, whether thinking is on, the input order, the instruction, and the output budget. Input order alone shifts evidence-following by up to 36.6 points. Such comparisons should also verify that answer length and alias matching do not drive the difference, as we do here for the crossed conflicts.

For systems that receive new evidence after reasoning, a separate answer call is the cheaper correction and renewed reasoning the stronger one. The distractor results of Section~\ref{sec:noise_boundary} make evidence quality part of this decision, and the stage-separated ablations favor improving retrieved evidence before the trace is written.

The two crossed directions partly separate the trace from what the model already knows. Trace-following is higher when the stale trace states the original answer, which points to parametric knowledge or to edits that still imply the original answer. Even so, continuation follows the stale trace in most conflicts in both directions.

\section{Related Work}
\label{sec:related}

\paragraph{Trace interventions and faithfulness metrics.}
Faithfulness studies probe stated reasoning with biasing features \citep{turpin2023language}, truncation and corruption \citep{lanham2023measuring}, causal mediation \citep{cmi2026}, and attention masking \citep{jia2026informationflow}, and examine unfaithful reasoning in the wild \citep{arcuschin2025chain}. FRODO pairs an original problem with reasoning generated for a modified problem to measure the influence of the reasoning chain \citep{paul2024frodo}. CASE evaluates perturbation-based and counterfactual metrics jointly and trains answer generation to depend on the trace \citep{case2026}. \citet{zaman2025causallens} compare the diagnostic performance of faithfulness metrics on explanation pairs constructed through model editing. We measure trace contribution and conflict response against verified current evidence on the same pairs and traces, and thereby show that, across answer configurations, the former does not reveal the latter.

\paragraph{Updated evidence and answer generation.}
Knowledge-conflict studies compare retrieved context with parametric memory \citep{longpre2021entity,xie2024adaptive}. Related interventions update context during math and code reasoning \citep{interruptible2025}, transplant trace prefixes generated under misleading cues \citep{linzmayer2026untangling}, and swap images after a vision-language model has reasoned \citep{shi2026visualswap}. The image-swap study shows that a new user turn can restore the use of a changed image while earlier reasoning is retained. Both of its compared conditions allow further reasoning. In our comparison, neither configuration reasons further, both answer from the same text traces and evidence pairs, and trace removal on these pairs measures how much each configuration uses the trace.

Appendix~\ref{app:related} covers further work on what traces carry and on evidence selection and use.

\section{Conclusion}
\label{sec:conclusion}

Trace use and conflict response are two measurable properties of final answer generation, and they diverge. On fixed model-generated traces and evidence pairs, gpt-oss-20b, Gemma4-E4B, and Qwen3-14B follow the current evidence 29.9 to 60.6 points more often under a separate answer call than under continuation, while removing the trace costs both configurations comparable accuracy. The value of a trace lies mostly in its final conclusion, evidence helps most before that conclusion is written, and renewed reasoning corrects a stale conclusion far more reliably than an instruction to verify it. These findings motivate evaluating trace contribution and conflict response jointly, with the answer configuration explicitly reported.

\newpage

\section*{Limitations}

Our paired comparisons cover three models from distinct families and two English multi-hop QA datasets. We study controlled conflicts constructed by editing evidence and retain pairs whose traces explicitly express the competing answers. The extent to which these findings generalize to naturally updated evidence, other tasks, and less explicit traces remains to be established.

\section*{Ethical Considerations}
We use public benchmarks and publicly released or commercially available models. The study involves no human subjects and no personal data.

\bibliography{mybib}

\appendix
\makeatletter
\setlength{\@fptop}{0pt}
\setlength{\@fpsep}{12pt plus 2pt}
\setlength{\@fpbot}{0pt plus 1fil}
\setlength{\@dblfptop}{0pt}
\setlength{\@dblfpsep}{10pt plus 2pt}
\setlength{\@dblfpbot}{0pt plus 1fil}
\makeatother
\section{Additional Related Work}
\label{app:related}

\paragraph{Information carried by traces.}
\citet{yu2026outcome} study whether answers can be recovered from reasoning under outcome-based training, and work on thinking drafts examines how appended conclusions affect final answers \citep{xiong2025drafts}. Other studies investigate traces as compressed context \citep{tac2026}, as task state \citep{traceasstate2026}, as transferable reasoning \citep{travels2026}, and as a lossy account of computation \citep{cotexplain2025}. Our trace-content controls complement these studies by quantifying how much selected facts and the conclusion of a trace each contribute to short-answer accuracy, and, for Qwen3-4B and Qwen3-14B, they show that the final line alone retains most of the contribution of a trace (Section~\ref{sec:follow}).

\paragraph{Evidence selection and use.}
IRCoT \citep{trivedi2023ircot} and SearChain \citep{xu2024searchain} combine retrieval with reasoning for multi-hop questions, and the position, relevance, and noise of the context affect how models use evidence \citep{liu2024lost,yoran2024making,noisybench2026,masking2025mhqa}. Related work analyzes evidence-location and integration failures \citep{zhang2026failure}, single-evidence interventions \citep{cuer2026}, context filtering \citep{wang2023filco,attentionrag2025,attrieval2025}, context attribution \citep{chuang2025selfcite}, and cache compression \citep{zhang2023h2o,li2024snapkv,speckv2025,accevidence2026}. Our stage-separated comparisons add a further distinction to this literature, namely whether a context intervention occurs before the trace is generated or after a trace already exists. Section~\ref{sec:enters} shows that it matters: with the same gold evidence at answer time, generating the trace from that evidence yields the larger gain.

\section{Experimental Details}
\label{app:setup}

\subsection{Conditions and Coverage}
\label{app:coverage}

\begin{table*}[t]
\centering
\small
\setlength{\tabcolsep}{2.5pt}
\begin{tabular}{lccc}
\toprule
Condition in the paper & Trace written from & Trace given & Answer context \\
\midrule
Trace + gold + distr., Full$\rightarrow$Full & Full & $r$ & Full \\
Gold + distr., no trace & Full & placeholder & Full \\
Half trace + gold + distr. & Full & about half of $r$ & Full \\
Trace only & Full & $r$ & none \\
Trace + distr., Full$\rightarrow$Distr. & Full & $r$ & Distr. \\
Trace + gold, Full$\rightarrow$Gold & Full & $r$ & Gold \\
Nothing & no trace & placeholder & none \\
Question-only trace and answer & question only & single call & single call \\
Gold$\rightarrow$Gold & Gold & $r$ & Gold \\
Distr.$\rightarrow$Distr. & Distr. & $r$ & Distr. \\
Gold$\rightarrow$Full & Gold & $r$ & Full \\
Gold$\rightarrow$Distr. & Gold & $r$ & Distr. \\
Distr.$\rightarrow$Gold & Distr. & $r$ & Gold \\
Distr.$\rightarrow$Full & Distr. & $r$ & Full \\
\midrule
Final line only & Full & last step of $r$ & Full \\
Trace minus final line & Full & $r$ without last step & Full \\
Extractive summary & gold sentences & length-matched & Full \\
Unrelated trace & another question & length-matched & Full \\
No trace (empty) & Full & empty & Full \\
Cache masking (Table~\ref{tab:kv_replication}) & Full & $r$, masked in the cache & masked in the cache \\
Crossed conflict $(E_i, r_j)$ & $E_j$ & $r_j$ & $E_i$ (alone or + distr.) \\
\bottomrule
\end{tabular}
\caption{Experimental conditions. For each condition, the columns specify the context from which the trace is written, the version of the trace supplied at answer time, and the answer-time context. Full, Gold, and Distr.\ denote the complete context, the annotated supporting paragraphs, and the distractor paragraphs, respectively. The upper block comprises the answer-time, reasoning-time, and cross-stage conditions, and the lower block the trace-content controls, the cache-masking implementation, and the crossed conflicts.}
\label{tab:conditions}
\end{table*}

Table~\ref{tab:conditions} defines every condition used in the paper. The main answer-time ablations cover the four Qwen3 sizes, Qwen3.5-4B, and Gemma4-E4B on both datasets (Table~\ref{tab:post}), and the reasoning-time ablations in Table~\ref{tab:pre} cover the same six Qwen and Gemma models on both datasets. The gpt-oss-20b runs additionally cover the answer-time, reasoning-time, and cross-stage conditions under both continuation and separate calls on both datasets. Gemma4-E4B is also run with continuation answers on the answer-time, reasoning-time, and cross-stage conditions of both datasets (Gemma4-E4B (Cont.\@) in Tables~\ref{tab:contrasts_hotpotqa} and~\ref{tab:contrasts_musique}). In these runs, the answer prompt repeats the reasoning prompt with the answer-time evidence, places the given trace in the model's thought channel, and closes the channel, after which the model generates the answer. Because these runs reuse the traces of the separate-call run, the two answer configurations are compared on identical traces. When the trace is empty, the thought channel is empty and the model reasons again before it answers. Every Gemma4-E4B (Cont.\@) contrast therefore uses only questions with non-empty traces. With the no-trace placeholder, 2.3\% (HotpotQA) and 15.1\% (MuSiQue) of its answers exceed 120 characters and are all scored as wrong, so its No trace $-$ full contrast partly reflects the answer format in addition to trace use.

Half trace removes $\max(1, \lfloor k/2 \rfloor)$ of the $k$ steps (non-empty lines) of a trace, chosen at random. Appendix~\ref{app:deepseek} describes the DeepSeek-V4-Pro runs.

The crossed conflicts cover the four Qwen3 sizes, gpt-oss-20b, and Gemma4-E4B on both datasets (Table~\ref{tab:conflict}, which also gives the retained pairs of each model), and DeepSeek-V4-Pro on HotpotQA (Appendix~\ref{app:deepseek}). The paired comparison of continuation and separate calls on the same evidence pairs and traces covers one model from each of three families: gpt-oss-20b, Gemma4-E4B, and Qwen3-14B.

\subsection{Decoding and Question Selection}

The reasoning stage instructs the model to reason over the supplied evidence (in numbered steps for Qwen3), and the answer stage requests a short final answer based on the trace and evidence. Section~\ref{sec:inference} describes the prompt structures of continuation and the separate call, and Appendix~\ref{app:prompts} reproduces the prompts.

\paragraph{No-trace placeholder and decoding.}
Conditions without a trace supply the text \texttt{[REASONING MASKED]} in place of the trace. All local models decode greedily, without sampling, so temperature and top-p do not apply, and gpt-oss-20b runs in vLLM with temperature 0. DeepSeek-V4-Pro uses the API defaults with low reasoning effort. Table~\ref{tab:decoding} lists the length limits. Apart from the MuSiQue context cut, no prompt is truncated.

\begin{table*}[t]
\centering
\footnotesize
\setlength{\tabcolsep}{3pt}
\begin{tabular}{lccc}
\toprule
Model & Thinking & Reasoning / answer tokens & Prompt limit \\
\midrule
Qwen3-4B/8B/14B/32B & prompted & 512 / 64 & 4,096 (H), 10,240 (M) \\
Qwen3.5-4B & native & 4,096 / 128 & 8,192 \\
Gemma4-E4B & native & 4,096 / 64 & 4,096 (H), 8,192 (M) \\
gpt-oss-20b & native, effort low & 4,096 / 64 & 8,192 (H), 10,240 (M) \\
DeepSeek-V4-Pro & native, effort low & 8,192 / 128 & API \\
\bottomrule
\end{tabular}
\caption{Generation limits for the reasoning and answer stages. Crossed-conflict runs use a prompt limit of at least 8,192 tokens.}
\label{tab:decoding}
\end{table*}

In the crossed conflicts with gold evidence alone, answers are short under both configurations, so answer length does not account for the configuration effect (Section~\ref{sec:components}). The median answer has 4 to 5 tokens under continuation and under the separate call alike (95th percentiles of 11 to 19 and 10 to 20 tokens, respectively), and at most 0.9\% of continuation answers continue reasoning in place of a short answer (counted when the answer starts like reasoning or is longer than 150 characters). Lengths are measured on the extracted answers: the text inside the answer tags for Qwen3, the first line of the final channel for gpt-oss-20b, and the reply after the thought channel for Gemma4-E4B.

HotpotQA uses the first 2,000 of the 7,405 validation questions of the distractor setting, in the order of the released file, together with the DeepSeek-V4-Pro subsets of Appendix~\ref{app:deepseek}. The main MuSiQue runs use the first 2,000 of the 2,417 validation questions of the answerable MuSiQue set, likewise in file order and without sampling, and cut the context at 12,000 characters (Section~\ref{sec:setup}). The first 2,000 questions of the MuSiQue file comprise 1,179 two-hop, 647 three-hop, and 174 four-hop questions, against 1,252, 760, and 405 in the full validation set.

\begin{table*}[t]
\centering
\footnotesize
\setlength{\tabcolsep}{3pt}
\begin{tabular}{lcccccc}
\toprule
& \multicolumn{3}{c}{EM (\%)} & \multicolumn{3}{c}{Paired difference (points)} \\
\cmidrule(lr){2-4}\cmidrule(lr){5-7}
& & & & Gold$\rightarrow$Gold & Full$\rightarrow$Gold & Gold$\rightarrow$Gold \\
Model & Full$\rightarrow$Full & Full$\rightarrow$Gold & Gold$\rightarrow$Gold & $-$ Full$\rightarrow$Full & $-$ Full$\rightarrow$Full & $-$ Full$\rightarrow$Gold \\
\midrule
Qwen3-4B & 31.2 & 34.4 & 45.7 & +14.5 [+12.4, +16.6] & +3.2 [+2.1, +4.4] & +11.3 [+9.3, +13.3] \\
Qwen3-8B & 36.6 & 41.1 & 47.8 & +11.2 [+9.0, +13.2] & +4.4 [+3.1, +5.7] & +6.7 [+4.9, +8.6] \\
Qwen3-14B & 37.9 & 39.1 & 49.0 & +11.0 [+9.0, +12.9] & +1.1 [+0.3, +2.0] & +9.9 [+7.8, +11.8] \\
Qwen3-32B & 40.6 & 47.6 & 50.0 & +9.3 [+7.5, +11.3] & +7.0 [+5.4, +8.6] & +2.4 [+0.9, +3.8] \\
Qwen3.5-4B & 40.6 & 46.3 & 50.0 & +9.3 [+7.2, +11.7] & +5.7 [+4.1, +7.5] & +3.7 [+1.9, +5.5] \\
Gemma4-E4B & 39.1 & 41.4 & 49.5 & +10.4 [+8.5, +12.4] & +2.4 [+1.5, +3.2] & +8.1 [+6.2, +10.0] \\
\bottomrule
\end{tabular}
\caption{MuSiQue stage-separated ablations restricted to the 1,829 of 2,000 questions whose cut Full context keeps every supporting paragraph (Section~\ref{sec:setup}). Full$\rightarrow$Full and Full$\rightarrow$Gold follow Table~\ref{tab:post}, and Gold$\rightarrow$Gold follows Table~\ref{tab:pre}. For every model, Gold$\rightarrow$Gold improves over both Full$\rightarrow$Full and Full$\rightarrow$Gold, so the advantage of generating the trace from gold evidence does not depend on the truncated questions. Paired differences have 95\% bootstrap intervals over questions (Appendix~\ref{app:metrics}), and all 18 differences have an exact McNemar $p < 0.05$.}
\label{tab:musique_cut}
\end{table*}

The 12,000-character cut removes at least one supporting paragraph from the Full context of 171 of the 2,000 MuSiQue questions (8.6\%), while gold-only contexts are never cut. Most of these questions require three or four hops. Their Full$\rightarrow$Full EM is 13.5 to 17.5, and Gold$\rightarrow$Gold exceeds it by 19.9 to 31.6 points, so they enlarge the MuSiQue gains over Full$\rightarrow$Full. To confirm that the conclusion of Section~\ref{sec:enters} does not depend on these questions, Table~\ref{tab:musique_cut} repeats the MuSiQue stage-separated ablations without them, and the conclusion holds: Gold$\rightarrow$Gold still improves EM over Full$\rightarrow$Full for all six models, by 9.3 to 14.5 points, and its gain exceeds that of Full$\rightarrow$Gold by 2.4 to 11.3 points, with every paired 95\% bootstrap interval excluding zero.

\subsection{Metrics and Statistical Summaries}
\label{app:metrics}

EM and F1 use SQuAD normalization, which lowercases text, removes punctuation and the articles \emph{a}, \emph{an}, and \emph{the}, and collapses whitespace. Contains checks whether the gold-answer tokens occur as a contiguous run among the lowercased alphanumeric tokens of an answer. Answer change rates compare normalized answers. Trace contribution is defined on both evaluation units: on the original questions, it is the negated no-trace-minus-full EM contrast of Tables~\ref{tab:contrasts_hotpotqa} and~\ref{tab:contrasts_musique}, and on the crossed-conflict pairs, it is matched-cell accuracy with the trace minus accuracy without it (Table~\ref{tab:joint}).

All paired comparisons align questions across conditions. Intervals are percentile intervals from 2,000 bootstrap resamples of questions with a fixed seed, and McNemar tests are exact two-sided binomial tests on the discordant questions. Differences reported in the text and tables are computed from unrounded accuracies and can therefore differ by 0.1 points from differences between rounded table entries. For the crossed conflicts, the resampling unit is instead the evidence pair, and each pair contributes its change in evidence-following averaged over its two crossed outputs (Table~\ref{tab:boot}).

\subsection{Trace Reuse and Cross-Stage Subsets}
\label{app:trace_audit}

The answer-time ablations, the crossed conflicts, and the cross-stage runs all reuse saved traces: each trace is written once and then used under every answer-time context. For Qwen3-4B and Qwen3-14B, the gold-written and distractor-written traces are those of the reasoning-time ablations (Table~\ref{tab:pre}). In Table~\ref{tab:full_q3_4b_14b}, the Gold$\rightarrow$Full, Gold$\rightarrow$Distr., Distr.$\rightarrow$Gold, and Distr.$\rightarrow$Full rows come from a second answer pass over these traces on all 2,000 questions, and the other rows come from the ablations of Tables~\ref{tab:post} and~\ref{tab:pre}. The second pass closely reproduces the first: its Gold$\rightarrow$Gold and Distr.$\rightarrow$Distr.\ answers are identical for 93.5\% to 98.7\% of the questions, and its EM differs by at most 0.2 points. For these two models, the cross-stage contrasts of Tables~\ref{tab:contrasts_hotpotqa} and~\ref{tab:contrasts_musique} and their rows of Table~\ref{tab:repair_distr} take every cross-stage cell from the second pass, including the Gold$\rightarrow$Gold and Distr.$\rightarrow$Distr.\ answers that define the baseline groups, while the answer-time and reasoning-time contrasts of the same two tables use the first. The other models take all their cross-stage cells from a single pass.

The question sets differ by table. The cross-stage rows of Table~\ref{tab:full_q3_4b_14b} cover all 2,000 questions. The other cross-stage analyses use the questions whose gold-written and distractor-written traces are both non-empty, because an answer call given an empty trace has nothing to be anchored to. The cross-stage contrasts of Tables~\ref{tab:contrasts_hotpotqa} and~\ref{tab:contrasts_musique} use these questions ($n$ in each row). Table~\ref{tab:repair_distr} uses the Distr.$\rightarrow$Distr.\ failures among the questions with both traces non-empty, for example 1,777 of the 1,823 failures of Qwen3-4B and 1,789 of the 1,790 failures of Qwen3-14B on HotpotQA. Table~\ref{tab:repair_full} uses the Full$\rightarrow$Full failures among the questions whose full-context and gold-written traces are both non-empty, except for the Qwen3-4B and Qwen3-14B rows, which use all 2,000 questions.

Each repair table defines its baseline failure set explicitly: Table~\ref{tab:repair_full} begins with Full$\rightarrow$Full failures, and Table~\ref{tab:repair_distr} with Distr.$\rightarrow$Distr.\ failures. Continuation and separate calls can produce different baseline failure sets. For gpt-oss-20b in Table~\ref{tab:repair_distr}, the HotpotQA counts are 1,602 and 1,595, and the MuSiQue counts are 1,864 and 1,872, and each repair percentage summarizes its own set. Appendix~\ref{app:repair} therefore also compares the two configurations on the questions that fail under both. By contrast, the crossed-conflict comparison in Table~\ref{tab:conflict} uses a shared set of evidence pairs and traces for the two answer configurations of each paired model.

\section{Counterfactual Data and Further Conflict Analyses}
\label{app:planned}

\subsection{Evidence Pairs and Trace Filtering}

For the crossed conflicts of Section~\ref{sec:interventions}, we select HotpotQA bridge questions and MuSiQue two-hop questions whose gold answer $a_0$ occurs in the gold paragraphs. A proposer model supplies a counterfactual answer $a_1$ of the same type, every occurrence of $a_0$ in the gold paragraphs is replaced with $a_1$, and a verifier model must answer each version with its own target. This construction yields 683 verified HotpotQA evidence pairs and 417 MuSiQue pairs before model-specific trace filtering. For each model, we then retain the pairs in which each trace states the target of its own evidence version and not the other target, and evaluate the matched cells $(E_i, r_i)$ and the crossed cells $(E_i, r_j)$ with $i \neq j$.

The retained pairs therefore depend on both traces stating their own target, so comparisons across models can involve different pairs. Within each gpt-oss-20b, Gemma4-E4B, and Qwen3-14B comparison, however, continuation and separate calls receive identical evidence pairs and traces.

\paragraph{Candidates and counterfactual proposals.}
We start from the first 2,000 validation questions of each dataset (answerable questions for MuSiQue). Of these, 1,612 HotpotQA bridge questions without yes/no answers and 1,179 MuSiQue two-hop questions are candidates, and 1,246 and 973 of them have an answer $a_0$ of at most six tokens that occurs in the gold paragraphs and not in the distractors. For each of these, gemma-3-27b-it is asked for a counterfactual answer $a_1$ of the same type (prompt in Appendix~\ref{app:prompts}), and every occurrence of $a_0$ in the gold paragraphs, titles included, is replaced with $a_1$. We drop proposals that are empty, longer than 80 characters, or whose tokens contain or are contained in those of $a_0$, proposals that already occur in the context, and edits after which $E_1$ still contains $a_0$, a dataset alias of $a_0$, or the last word of a capitalized two- or three-word $a_0$ when that word has at least four letters and does not occur in $a_1$. This leaves 942 HotpotQA and 674 MuSiQue edited pairs.

\paragraph{Verification.}
The verifier is the same gemma-3-27b-it, an earlier generation of the Gemma family than Gemma4-E4B, with greedy decoding and at most 32 new tokens. It answers each version in its own request, which sees only the gold paragraphs of that version, each rendered as its title followed by its text, with the prompt in Appendix~\ref{app:prompts}. A pair passes if the answer for $E_0$ matches $a_0$ or, for MuSiQue, an alias of $a_0$, and the answer for $E_1$ matches $a_1$ and not $a_0$. Matching uses EM or Contains. Verification keeps 683 HotpotQA and 417 MuSiQue pairs.

\paragraph{Trace filtering.}
Trace filtering retains 382 to 642 of the 683 HotpotQA pairs and 281 to 364 of the 417 MuSiQue pairs, depending on the model (Table~\ref{tab:conflict}). A removed pair usually has a trace that does not state its own answer or, mostly for Gemma4-E4B and Qwen3-32B, an empty trace.

\paragraph{Residual leakage.}
The pairs are screened only automatically, so an edited version can still imply $a_0$ in ways that the string checks do not catch, for example through a translation, a partial name, or an unchanged fact. Such a leak can make an $a_0$ answer in $(E_1, r_0)$ resemble trace-following, but $(E_0, r_1)$ does not use $E_1$ as evidence, so such a leak cannot make its answers resemble trace-following. We therefore report the two directions separately below.

\paragraph{Results by direction.}
Table~\ref{tab:conflict_dir} splits the crossed cells of Table~\ref{tab:conflict} by direction. In $(E_0, r_1)$, the evidence is unedited and the stale trace states the counterfactual $a_1$, contrary to both the evidence and what the model may know. In $(E_1, r_0)$, the stale trace states the original answer $a_0$. Trace-following is higher in $(E_1, r_0)$ in every row. The margin is 4.5 to 25.7 points on HotpotQA and 4.1 to 16.2 points on MuSiQue, apart from Qwen3-4B and gpt-oss-20b with continuation on MuSiQue, where it is two answers (0.6 and 0.7 points). Knowledge of $a_0$ and edited evidence that still implies $a_0$ can both raise trace-following in $(E_1, r_0)$. The difference between the answer configurations holds in both directions, so neither source of asymmetry explains it: for gpt-oss-20b and Gemma4-E4B on both datasets, a separate call raises evidence-following over continuation by 52.7 to 62.1 points in $(E_0, r_1)$ and by 45.1 to 59.0 points in $(E_1, r_0)$, and for Qwen3-14B by 34.6 to 54.2 points and by 25.3 to 37.8 points, respectively.

\paragraph{Alias matching.}
Alias matching affects these results only marginally. Only MuSiQue provides aliases, and they are accepted only for $a_0$, which is the answer of the trace in $(E_1, r_0)$ and of the evidence in $(E_0, r_1)$. Without aliases, trace-following in $(E_1, r_0)$ falls by at most 1.8 points (Qwen3-8B, 5 of 281 answers), evidence-following in $(E_0, r_1)$ by at most 1.1 points (Qwen3-14B with a separate call, 4 of 364 answers), and matched accuracy by at most 0.9 points, while the other two crossed shares do not change. Alias matching therefore favors trace-following in $(E_1, r_0)$ by at most 1.8 points, which reverses the order of the two directions only for Qwen3-4B on MuSiQue: without aliases, Qwen3-4B follows the trace in 81.8\% of $(E_1, r_0)$ answers and 82.4\% of $(E_0, r_1)$ answers.

\begin{table}[t]
\centering
\footnotesize
\setlength{\tabcolsep}{2.6pt}
\resizebox{\columnwidth}{!}{%
\begin{tabular}{llcccccccc}
\toprule
& & \multicolumn{4}{c}{HotpotQA} & \multicolumn{4}{c}{MuSiQue} \\
\cmidrule(lr){3-6}\cmidrule(lr){7-10}
& & \multicolumn{2}{c}{$(E_0, r_1)$} & \multicolumn{2}{c}{$(E_1, r_0)$} & \multicolumn{2}{c}{$(E_0, r_1)$} & \multicolumn{2}{c}{$(E_1, r_0)$} \\
\cmidrule(lr){3-4}\cmidrule(lr){5-6}\cmidrule(lr){7-8}\cmidrule(lr){9-10}
Model & Config. & Trace & Evid. & Trace & Evid. & Trace & Evid. & Trace & Evid. \\
\midrule
Qwen3-4B & Cont. & 75.1 & 12.8 & 79.6 & 6.7 & 82.4 & 6.2 & 82.9 & 4.2 \\
Qwen3-8B & Cont. & 66.8 & 19.1 & 78.0 & 7.6 & 84.7 & 8.2 & 89.3 & 2.8 \\
Qwen3-14B & Cont. & 69.0 & 18.5 & 75.4 & 9.7 & 77.7 & 12.4 & 81.9 & 9.1 \\
Qwen3-14B & Sep. & 21.2 & 72.7 & 46.9 & 47.5 & 44.2 & 47.0 & 60.4 & 34.3 \\
Qwen3-32B & Cont. & 60.1 & 26.6 & 71.1 & 16.4 & 68.8 & 23.9 & 79.0 & 13.7 \\
gpt-oss-20b & Cont. & 61.7 & 31.7 & 77.1 & 16.3 & 82.3 & 12.3 & 82.9 & 11.3 \\
gpt-oss-20b & Sep. & 9.4 & 84.4 & 22.5 & 71.7 & 24.9 & 67.9 & 37.2 & 56.3 \\
Gemma4-E4B & Cont. & 70.6 & 19.8 & 84.5 & 7.4 & 71.6 & 13.2 & 80.0 & 5.8 \\
Gemma4-E4B & Sep. & 11.8 & 81.9 & 27.9 & 66.4 & 20.3 & 67.1 & 35.2 & 51.9 \\
\midrule
\multicolumn{10}{l}{\textit{Corrections, Qwen3-14B}} \\
Verify & Cont. & 70.4 & 16.8 & 76.8 & 9.1 & 77.5 & 12.1 & 83.2 & 7.4 \\
Re-reason & Fresh & 0.6 & 89.8 & 8.2 & 83.4 & 6.0 & 84.9 & 15.7 & 76.6 \\
\bottomrule
\end{tabular}%
}
\caption{Crossed cells of Table~\ref{tab:conflict} by direction, with gold evidence only and the same pairs, traces, and matching rules. Values are percentages of the crossed answers in each direction, and Other (not shown) is the remainder. In $(E_0, r_1)$ the evidence is unedited and the trace states the counterfactual $a_1$. In $(E_1, r_0)$ the evidence is edited and the trace states the original answer $a_0$. Averaging the two directions gives the crossed columns of Table~\ref{tab:conflict}. The correction rows cover both datasets, as in Table~\ref{tab:conflict}, and Fresh marks a new reasoning call.}
\label{tab:conflict_dir}
\end{table}

\subsection{Trace Use and Conflict Response on the Same Pairs}
\label{app:joint}

\begin{table*}[t]
\centering
\footnotesize
\setlength{\tabcolsep}{3pt}
\begin{tabular}{llccccccccccc}
\toprule
& & & \multicolumn{5}{c}{Gold evidence only} & \multicolumn{5}{c}{Gold evidence and distractors} \\
\cmidrule(lr){4-8}\cmidrule(lr){9-13}
Model & Config. & Pairs & Matched & No trace & Contrib. & Change & Evid. & Matched & No trace & Contrib. & Change & Evid. \\
\midrule
\multicolumn{13}{l}{\textit{HotpotQA}} \\
gpt-oss-20b & Cont. & 533 & 96.7 & 92.0 & 4.7 & 19.7 & 24.0 & 96.5 & 81.8 & 14.7 & 31.2 & 15.3 \\
gpt-oss-20b & Sep. & 533 & 96.2 & 91.6 & 4.7 & 22.0 & 78.0 & 96.0 & 84.9 & 11.1 & 30.0 & 71.0 \\
Gemma4-E4B & Cont. & 459 & 93.4 & 88.1 & 5.2 & 19.2 & 13.6 & 93.9 & 71.4 & 22.5 & 41.5 & 15.1 \\
Gemma4-E4B & Sep. & 459 & 93.4 & 85.1 & 8.3 & 19.6 & 74.2 & 93.2 & 70.7 & 22.5 & 34.9 & 73.9 \\
Qwen3-14B & Cont. & 638 & 87.1 & 80.6 & 6.6 & 26.7 & 14.1 & 87.0 & 67.2 & 19.7 & 36.0 & 17.6 \\
Qwen3-14B & Sep. & 638 & 93.7 & 90.0 & 3.7 & 22.8 & 60.1 & 92.9 & 80.5 & 12.5 & 35.1 & 45.5 \\
\midrule
\multicolumn{13}{l}{\textit{MuSiQue}} \\
gpt-oss-20b & Cont. & 293 & 96.4 & 82.8 & 13.7 & 31.7 & 11.8 & 95.6 & 57.5 & 38.1 & 59.4 & 7.0 \\
gpt-oss-20b & Sep. & 293 & 94.5 & 83.4 & 11.1 & 34.0 & 62.1 & 95.4 & 58.9 & 36.5 & 59.4 & 41.3 \\
Gemma4-E4B & Cont. & 310 & 87.7 & 73.4 & 14.4 & 39.7 & 9.5 & 86.9 & 41.5 & 45.5 & 71.6 & 7.7 \\
Gemma4-E4B & Sep. & 310 & 87.7 & 68.7 & 19.0 & 36.8 & 59.5 & 85.0 & 39.5 & 45.5 & 61.9 & 52.7 \\
Qwen3-14B & Cont. & 364 & 90.2 & 76.1 & 14.1 & 33.8 & 10.7 & 89.7 & 52.1 & 37.6 & 54.9 & 10.7 \\
Qwen3-14B & Sep. & 364 & 93.7 & 84.1 & 9.6 & 37.9 & 40.7 & 93.1 & 58.7 & 34.5 & 63.5 & 25.3 \\
\bottomrule
\end{tabular}
\caption{Joint measurement of trace use and conflict response on the same retained pairs and traces. Matched is the accuracy (\%) of the matched cells $(E_i,r_i)$ and repeats Tables~\ref{tab:conflict} and~\ref{tab:conflict_full}. No trace is the accuracy (\%) of $E_i$ answered with the no-trace placeholder, Contrib.\ is Matched minus No trace in points, and Change is the share (\%) of matched-cell answers that change when the trace is removed. Evid.\ is evidence-following (\%) in the crossed cells $(E_i,r_j)$, $i\neq j$.}
\label{tab:joint}
\end{table*}

Because the deletion ablations and the crossed conflicts use different questions and contexts, the joint evaluation of Section~\ref{sec:two_dimensions} measures trace contribution and conflict response on the same retained pairs and traces. For each pair and answer configuration, it adds a no-trace cell for each evidence version to the two matched cells $(E_i,r_i)$ and the two crossed cells $(E_i,r_j)$, $i\neq j$. A no-trace cell keeps the prompt of its answer configuration and supplies the no-trace placeholder of the main ablations in place of the trace (Table~\ref{tab:conditions}). Each pair thus comprises six cells per configuration and twelve across continuation and the separate call.

The runs use gpt-oss-20b, Gemma4-E4B, and Qwen3-14B on both datasets, each with gold evidence alone and with the original distractors, and reuse the stored traces and retained pairs of Table~\ref{tab:conflict}. The separate call of Qwen3-14B was added for this analysis. Table~\ref{tab:joint} reports, for each model, configuration, and context, the accuracy of the matched cells with and without the trace, their difference as the trace contribution, the share of matched-cell answers that change when the trace is removed, and evidence-following in the crossed cells. Differences between configurations carry pair-level bootstrap intervals as in Table~\ref{tab:boot}.

On the same pairs, the two measurements diverge. Removing the trace from the matched cells costs 3.7 to 8.3 points with gold evidence alone on HotpotQA and 9.6 to 19.0 points on MuSiQue. The trace contribution of continuation minus that of the separate call is $+0.0$ [$-1.5$, $+1.5$] for gpt-oss-20b, $-3.1$ [$-5.8$, $-0.4$] for Gemma4-E4B, and $+2.9$ [$+0.5$, $+5.3$] for Qwen3-14B on HotpotQA, and $+2.6$ [$+0.0$, $+5.5$], $-4.7$ [$-9.0$, $-0.8$], and $+4.5$ [$+0.0$, $+8.5$] on MuSiQue. The Gemma4-E4B intervals and the Qwen3-14B HotpotQA interval exclude zero, so some of these differences are statistically detectable. However, their sign varies across models, and none exceeds 4.7 points in magnitude. In the crossed cells of the same pairs, by contrast, continuation follows the evidence 46.0 to 60.6 points less often than the separate call on HotpotQA and 29.9 to 50.3 points less often on MuSiQue. The trace contributions of the two configurations thus differ by at most 4.7 points, while their evidence-following differs by 29.9 to 60.6 points on the same pairs. The separate call of Qwen3-14B follows the evidence in 60.1\% (HotpotQA) and 40.7\% (MuSiQue) of crossed cells, against 14.1\% and 10.7\% with continuation, so the effect of the answer configuration also holds for a third model family. With distractors added, the trace contributes more (11.1 to 22.5 points on HotpotQA, 34.5 to 45.5 on MuSiQue), and the same pattern holds (Table~\ref{tab:joint}).

\subsection{Changing One Part of the Answer Configuration}
\label{app:components}

Continuation and the separate call differ in three respects simultaneously (Section~\ref{sec:inference}). In continuation, the trace sits in the model's own thinking block after the evidence, and the reasoning instruction stays in place. In the separate call, the trace is user-supplied text, the evidence comes last, and a new instruction asks for a short answer. To attribute the configuration effect to these factors, the component runs add variants of the separate call, each of which changes a single element while keeping all other text identical. The first lists the evidence before the trace, so that the trace comes last as in continuation. The second keeps the order of the separate call and labels the trace as the model's own earlier reasoning for this question. Table~\ref{tab:components} orders four configurations so that neighboring columns differ in trace placement and instruction (first and second), input order (second and third), and the provenance label (third and fourth).

The runs reuse the traces and retained pairs of Table~\ref{tab:conflict} for gpt-oss-20b, Gemma4-E4B, and Qwen3-14B on both datasets, with gold evidence alone and with the original distractors. Evidence-following is computed as in Table~\ref{tab:conflict}, and each variant is compared with the separate call by pair-level bootstrap intervals as in Table~\ref{tab:boot}.

Moving the trace after the evidence lowers evidence-following the most: relative to the separate call, it falls by 7.2 [4.6, 9.9], 36.4 [32.4, 40.6], and 31.3 [28.0, 34.8] points for gpt-oss-20b, Gemma4-E4B, and Qwen3-14B on HotpotQA, and by 20.3, 36.6, and 27.2 points on MuSiQue (Table~\ref{tab:components}). For Gemma4-E4B and Qwen3-14B, this single change closes 60\% to 91\% of the gap between the separate call and continuation. For gpt-oss-20b, most of the gap remains, so placing the trace in the thinking block together with the reasoning instruction matters more for this model. Labeling the trace as the model's own earlier reasoning does not increase reliance on it. Evidence-following instead rises by 3.2 to 7.7 points. With the original distractors added, the order effect is 12.5 to 29.4 points on HotpotQA and 9.8 to 26.0 points on MuSiQue, and the own-label effect is $+1.4$ to $+9.1$ and $+5.5$ to $+8.1$ points. A third variant that labels the trace as written by another model changes evidence-following by $-0.9$ to $+8.2$ points on HotpotQA and $-1.4$ to $+10.2$ on MuSiQue relative to the separate call.

\begin{table}[t]
\centering
\footnotesize
\setlength{\tabcolsep}{3pt}
\begin{tabular}{lcccc}
\toprule
& Cont. & Sep. & Sep. & Sep. \\
& & evid. first & & own label \\
\midrule
Trace in & thinking & user turn & user turn & user turn \\
Instruction & reasoning & answer & answer & answer \\
Last input & trace & trace & evidence & evidence \\
Own label & implied & no & no & yes \\
\midrule
\multicolumn{5}{l}{\textit{HotpotQA}} \\
gpt-oss-20b & 24.0 & 70.8 & 78.0 & 81.2 \\
Gemma4-E4B & 13.6 & 37.8 & 74.2 & 78.0 \\
Qwen3-14B & 14.1 & 28.8 & 60.1 & 66.2 \\
\midrule
\multicolumn{5}{l}{\textit{MuSiQue}} \\
gpt-oss-20b & 11.8 & 41.8 & 62.1 & 66.4 \\
Gemma4-E4B & 9.5 & 22.9 & 59.5 & 66.5 \\
Qwen3-14B & 10.7 & 13.5 & 40.7 & 48.4 \\
\bottomrule
\end{tabular}
\caption{Component comparison: evidence-following (\%) in the crossed cells with gold evidence only, on the same traces and retained pairs as Table~\ref{tab:conflict}. The header rows give where the trace sits, which instruction is given, which input comes last, and whether the trace is labeled as the model's own, which the thinking block implies in continuation.}
\label{tab:components}
\end{table}

\section{Full Results}
\label{app:full}

Cont.\ and Sep.\ denote continuation and separate answer calls.

\subsection{Crossed Conflicts Between Trace and Evidence}
\label{app:conflicts}

Tables~\ref{tab:conflict} and~\ref{tab:conflict_full} report matched-cell accuracy and the distribution of crossed-cell answers with gold evidence alone and with the original distractors added, respectively. The pair counts after model-specific trace filtering appear in Table~\ref{tab:conflict}, and Appendix~\ref{app:planned} describes the construction and validation of the pairs. The last two rows of Table~\ref{tab:conflict} repeat the correction procedures of Section~\ref{sec:whichpart}: Verify modifies the continuation instruction, and Re-reason adds a new reasoning call. The first-pass reference of 90.2\% in that section is the share of the 1,276 first reasoning calls on both evidence versions of these 638 pairs whose answer matches the target of their evidence. Table~\ref{tab:boot} gives paired bootstrap intervals for the differences between answer configurations.

Model scale raises continuation evidence-following only modestly. On the 311 HotpotQA and 207 MuSiQue pairs that all four Qwen3 sizes retain, continuation evidence-following is 8.7\%, 14.0\%, 13.5\%, and 21.4\% on HotpotQA and 4.3\%, 5.8\%, 9.9\%, and 18.6\% on MuSiQue for 4B, 8B, 14B, and 32B, each within 1.1 points of the rate on that size's own retained set in Table~\ref{tab:conflict}. The 32B minus 4B differences are $+12.7$ points [$+9.2$, $+16.1$] on HotpotQA and $+14.3$ points [$+10.1$, $+18.4$] on MuSiQue, with 95\% bootstrap intervals over evidence pairs. These increases are less than half of the $+46.0$ and $+29.9$ points that a separate call yields for Qwen3-14B on its retained pairs (Table~\ref{tab:boot}), and even Qwen3-32B follows the stale trace in most of its crossed answers (Table~\ref{tab:conflict}). Of the differences between adjacent sizes, only 14B minus 8B on HotpotQA ($-0.5$ points [$-3.4$, $+2.3$]) and 8B minus 4B on MuSiQue ($+1.4$ points [$-1.2$, $+3.9$]) have intervals that include zero.

\begin{table*}[t]
\centering
\footnotesize
\setlength{\tabcolsep}{3.5pt}
\begin{tabular}{llcccccccccc}
\toprule
& & \multicolumn{5}{c}{HotpotQA} & \multicolumn{5}{c}{MuSiQue} \\
\cmidrule(lr){3-7}\cmidrule(lr){8-12}
& & & Matched & \multicolumn{3}{c}{Stale trace, $i \neq j$ (\%)} & & Matched & \multicolumn{3}{c}{Stale trace, $i \neq j$ (\%)} \\
\cmidrule(lr){5-7}\cmidrule(lr){10-12}
Model & Config. & Pairs & (\%) & Trace & Evid. & Other & Pairs & (\%) & Trace & Evid. & Other \\
\midrule
Qwen3-4B & Cont. & 642 & 87.1 & 77.3 & 9.7 & 12.9 & 357 & 88.8 & 82.6 & 5.2 & 12.2 \\
Qwen3-8B & Cont. & 382 & 88.2 & 72.4 & 13.4 & 14.3 & 281 & 92.2 & 87.0 & 5.5 & 7.5 \\
Qwen3-14B & Cont. & 638 & 87.1 & 72.2 & 14.1 & 13.7 & 364 & 90.2 & 79.8 & 10.7 & 9.5 \\
Qwen3-14B & Sep. & 638 & 93.7 & 34.0 & 60.1 & 5.9 & 364 & 93.7 & 52.3 & 40.7 & 7.0 \\
Qwen3-32B & Cont. & 561 & 88.4 & 65.6 & 21.5 & 12.9 & 314 & 91.7 & 73.9 & 18.8 & 7.3 \\
gpt-oss-20b & Cont. & 533 & 96.7 & 69.4 & 24.0 & 6.6 & 293 & 96.4 & 82.6 & 11.8 & 5.6 \\
gpt-oss-20b & Sep. & 533 & 96.2 & 15.9 & 78.0 & 6.0 & 293 & 94.5 & 31.1 & 62.1 & 6.8 \\
Gemma4-E4B & Cont. & 459 & 93.4 & 77.6 & 13.6 & 8.8 & 310 & 87.7 & 75.8 & 9.5 & 14.7 \\
Gemma4-E4B & Sep. & 459 & 93.4 & 19.8 & 74.2 & 6.0 & 310 & 87.7 & 27.7 & 59.5 & 12.7 \\
\midrule
\multicolumn{12}{l}{\textit{Corrections, Qwen3-14B on HotpotQA}} \\
Verify & Cont. & 638 & 87.1 & 73.6 & 12.9 & 13.5 & 364 & 90.8 & 80.4 & 9.8 & 9.9 \\
Re-reason & Fresh & 638 & 90.1 & 4.4 & 86.6 & 9.0 & 364 & 91.5 & 10.9 & 80.8 & 8.4 \\
\bottomrule
\end{tabular}
\caption{Crossed conflicts with gold evidence only (Figure~\ref{fig:conflict}). Pairs are verified evidence pairs whose two traces each state their own answer (Appendix~\ref{app:planned}). Matched is the share of answers in cells $(E_i,r_i)$ that match $a_i$. For crossed cells $(E_i,r_j)$ with $i\neq j$, Trace matches only $a_j$, Evid.\ matches only $a_i$, and Other matches neither or both. Matching uses exact match or Contains, and answers can also match dataset aliases of the original answer $a_0$. In the Config.\ column, Fresh marks a new reasoning call.}
\label{tab:conflict}
\end{table*}

\begin{table*}[t]
\centering
\footnotesize
\setlength{\tabcolsep}{4pt}
\begin{tabular}{llcccccccc}
\toprule
& & \multicolumn{4}{c}{HotpotQA} & \multicolumn{4}{c}{MuSiQue} \\
\cmidrule(lr){3-6}\cmidrule(lr){7-10}
& & Matched & \multicolumn{3}{c}{Stale trace, $i \neq j$} & Matched & \multicolumn{3}{c}{Stale trace, $i \neq j$} \\
\cmidrule(lr){4-6}\cmidrule(lr){8-10}
Model & Config. & (\%) & Trace & Evid. & Other & (\%) & Trace & Evid. & Other \\
\midrule
Qwen3-4B & Cont. & 86.4 & 72.8 & 12.9 & 14.3 & 87.1 & 78.3 & 7.1 & 14.6 \\
Qwen3-8B & Cont. & 87.4 & 71.1 & 14.5 & 14.4 & 92.2 & 83.5 & 5.9 & 10.7 \\
Qwen3-14B & Cont. & 87.0 & 66.7 & 17.6 & 15.7 & 89.7 & 76.8 & 10.7 & 12.5 \\
Qwen3-14B & Sep. & 92.9 & 48.7 & 45.5 & 5.8 & 93.1 & 67.2 & 25.3 & 7.6 \\
Qwen3-32B & Cont. & 87.1 & 65.8 & 21.9 & 12.3 & 92.4 & 76.6 & 15.0 & 8.4 \\
gpt-oss-20b & Cont. & 96.5 & 78.2 & 15.3 & 6.5 & 95.6 & 87.5 & 7.0 & 5.5 \\
gpt-oss-20b & Sep. & 96.0 & 23.2 & 71.0 & 5.8 & 95.4 & 52.2 & 41.3 & 6.5 \\
Gemma4-E4B & Cont. & 93.9 & 76.9 & 15.1 & 8.0 & 86.9 & 76.6 & 7.7 & 15.6 \\
Gemma4-E4B & Sep. & 93.2 & 19.6 & 73.9 & 6.5 & 85.0 & 29.0 & 52.7 & 18.2 \\
\bottomrule
\end{tabular}
\caption{Crossed conflicts with the original distractors added to each gold-evidence version, using the same pairs and traces as Table~\ref{tab:conflict}. Values are percentages, and columns and matching rules follow Table~\ref{tab:conflict}.}
\label{tab:conflict_full}
\end{table*}

\begin{table*}[t]
\centering
\footnotesize
\setlength{\tabcolsep}{4pt}
\begin{tabular}{lcccccccc}
\toprule
& \multicolumn{4}{c}{HotpotQA} & \multicolumn{4}{c}{MuSiQue} \\
\cmidrule(lr){2-5}\cmidrule(lr){6-9}
Model & Pairs & Cont. & Sep. & Sep. $-$ Cont. & Pairs & Cont. & Sep. & Sep. $-$ Cont. \\
\midrule
gpt-oss-20b & 533 & 24.0 & 78.0 & $+54.0\,[+50.6,+57.4]$ & 293 & 11.8 & 62.1 & $+50.3\,[+45.7,+55.3]$ \\
Gemma4-E4B & 459 & 13.6 & 74.2 & $+60.6\,[+56.8,+64.5]$ & 310 & 9.5 & 59.5 & $+50.0\,[+45.2,+55.3]$ \\
Qwen3-14B & 638 & 14.1 & 60.1 & $+46.0\,[+42.8,+49.3]$ & 364 & 10.7 & 40.7 & $+29.9\,[+25.3,+34.6]$ \\
\bottomrule
\end{tabular}
\caption{Evidence-following (\%) of continuation and a separate call on the same crossed-conflict pairs and traces with gold evidence only (Table~\ref{tab:conflict}), and their paired difference in points with a 95\% pair-level bootstrap interval (Appendix~\ref{app:metrics}).}
\label{tab:boot}
\end{table*}

\subsection{Evidence Removal Before and After Reasoning}

Tables~\ref{tab:post} and~\ref{tab:pre} report EM for the answer-time and reasoning-time ablations summarized in Figure~\ref{fig:ablations}, and Table~\ref{tab:change} reports answer changes, along with empty-trace frequencies that inform the interpretation of trace removal for models that often return empty traces (Section~\ref{sec:follow}).

\begin{table*}[t]
\centering
\footnotesize
\setlength{\tabcolsep}{4pt}
\begin{tabular}{lcccccccccccc}
\toprule
& \multicolumn{6}{c}{HotpotQA} & \multicolumn{6}{c}{MuSiQue} \\
\cmidrule(lr){2-7}\cmidrule(lr){8-13}
& \multicolumn{4}{c}{Qwen3} & Qwen3.5 & Gemma4 & \multicolumn{4}{c}{Qwen3} & Qwen3.5 & Gemma4 \\
\cmidrule(lr){2-5}\cmidrule(lr){8-11}
Answer input & 4B & 8B & 14B & 32B & 4B & E4B & 4B & 8B & 14B & 32B & 4B & E4B \\
\midrule
Trace + gold + distr. & 48.0 & 50.9 & 53.2 & 53.2 & 61.8 & 58.6 & 29.6 & 35.0 & 35.9 & 38.6 & 38.5 & 37.2 \\
\midrule
Gold + distr., no trace & 41.5 & 46.6 & 43.6 & 47.4 & 54.3 & 49.5 & 18.1 & 22.2 & 25.5 & 28.9 & 27.5 & 21.8 \\
Half trace + gold + distr. & 46.8 & 50.3 & 50.5 & 50.9 & 62.5 & 58.3 & 27.7 & 32.4 & 34.1 & 37.1 & 39.6 & 36.4 \\
Trace only & 50.7 & 37.6 & 50.6 & 49.8 & 56.2 & 57.0 & 30.3 & 29.3 & 33.5 & 29.5 & 25.4 & 31.7 \\
Trace + distr. & 44.7 & 35.6 & 49.5 & 47.7 & 56.8 & 52.9 & 27.0 & 27.9 & 33.1 & 28.9 & 27.6 & 33.6 \\
Trace + gold & 50.2 & 52.8 & 53.6 & 53.2 & 63.8 & 58.6 & 33.2 & 39.5 & 37.2 & 46.6 & 44.8 & 40.6 \\
Nothing & 13.3 & 14.8 & 16.1 & 17.7 & 17.9 & 16.1 & 1.8 & 2.6 & 3.5 & 4.0 & 3.6 & 2.6 \\
\midrule
\textit{Empty trace (\%)} & 0.3 & 0.1 & 0.1 & 8.2 & 21.0 & 2.2 & 3.6 & 1.7 & 0.3 & 38.9 & 61.2 & 0.3 \\
\textit{Gold answer in trace (\%)} & 77.3 & 49.2 & 77.5 & 78.4 & 75.4 & 89.6 & 54.2 & 47.6 & 59.9 & 51.3 & 36.2 & 71.0 \\
\bottomrule
\end{tabular}
\caption{Answer-time ablations. The trace is always written from the full context, and only the answer-stage input changes. Values are EM (\%) on 2,000 questions per dataset. The last two rows describe the full-context traces: the share that is empty and the share that contains the gold answer.}
\label{tab:post}
\end{table*}

\begin{table*}[t]
\centering
\small
\setlength{\tabcolsep}{3.6pt}
\begin{tabular}{lcccccccccccc}
\toprule
& \multicolumn{6}{c}{HotpotQA} & \multicolumn{6}{c}{MuSiQue} \\
\cmidrule(lr){2-7}\cmidrule(lr){8-13}
& \multicolumn{4}{c}{Qwen3} & Qwen3.5 & Gemma4 & \multicolumn{4}{c}{Qwen3} & Qwen3.5 & Gemma4 \\
\cmidrule(lr){2-5}\cmidrule(lr){8-11}
Context & 4B & 8B & 14B & 32B & 4B & E4B & 4B & 8B & 14B & 32B & 4B & E4B \\
\midrule
Full context & 48.0 & 50.9 & 53.2 & 53.2 & 61.8 & 58.6 & 29.6 & 35.0 & 35.9 & 38.6 & 38.5 & 37.2 \\
Gold only & 54.0 & 55.7 & 56.0 & 56.0 & 64.7 & 62.1 & 44.6 & 47.0 & 48.3 & 49.5 & 49.1 & 49.5 \\
Distractors only & 8.6 & 11.8 & 10.4 & 14.6 & 15.0 & 7.7 & 1.9 & 3.3 & 2.7 & 3.2 & 2.6 & 1.2 \\
Question only & 11.2 & 14.2 & 14.0 & 13.2 & 14.5 & 13.9 & 3.0 & 2.9 & 3.9 & 3.7 & 2.6 & 2.7 \\
\midrule
\multicolumn{13}{l}{\textit{Gold answer contained in the trace (\%)}} \\
Trace from full context & 77.3 & 49.2 & 77.5 & 78.4 & 75.4 & 89.6 & 54.2 & 47.6 & 59.9 & 51.3 & 36.2 & 71.0 \\
Trace from gold only & 80.5 & 53.4 & 80.0 & 83.4 & 80.3 & 72.6 & 70.7 & 63.0 & 75.5 & 66.9 & 54.7 & 80.5 \\
Trace from distractors only & 21.3 & 17.0 & 22.4 & 16.3 & 9.8 & 23.0 & 4.5 & 4.8 & 5.7 & 3.7 & 1.0 & 4.4 \\
\bottomrule
\end{tabular}
\caption{Reasoning-time ablations. Both stages use the context in the first column (one call for Question only), so the trace is written from that context. The first block gives EM (\%) on 2,000 questions per dataset, and Gold only yields the highest EM for every model. The second block gives the share of traces (\%) that contain the gold answer string, counting empty traces as not containing it.}
\label{tab:pre}
\end{table*}

\begin{table*}[t]
\centering
\footnotesize
\setlength{\tabcolsep}{4pt}
\begin{tabular}{lcccccccc}
\toprule
& \multicolumn{4}{c}{HotpotQA} & \multicolumn{4}{c}{MuSiQue} \\
\cmidrule(lr){2-5}\cmidrule(lr){6-9}
& \multicolumn{3}{c}{Answers changed (\%)} & Empty & \multicolumn{3}{c}{Answers changed (\%)} & Empty \\
\cmidrule(lr){2-4}\cmidrule(lr){6-8}
Model & $-$trace & $-$evidence & $-$distr. & trace & $-$trace & $-$evidence & $-$distr. & trace \\
\midrule
Qwen3-4B & 44.8 & 24.0 & 20.3 & 0.3 & 71.3 & 33.7 & 30.9 & 3.6 \\
Qwen3-8B & 36.2 & 41.6 & 20.1 & 0.1 & 62.4 & 36.4 & 27.0 & 1.7 \\
Qwen3-14B & 37.6 & 17.3 & 12.6 & 0.1 & 64.6 & 27.5 & 22.3 & 0.3 \\
Qwen3-32B & 31.8 & 24.6 & 17.6 & 8.2 & 39.7 & 48.4 & 36.4 & 38.9 \\
Qwen3.5-4B & 19.2 & 22.2 & 11.8 & 21.0 & 21.2 & 58.6 & 36.8 & 61.2 \\
Gemma4-E4B & 32.3 & 14.8 & 7.7 & 2.2 & 67.6 & 39.4 & 24.3 & 0.3 \\
\bottomrule
\end{tabular}
\caption{Share of answers that change when the answer stage loses the trace ($-$trace), all evidence ($-$evidence), or only the distractors ($-$distr.), relative to standard inference with the same trace, on 2,000 questions per dataset. The Empty trace columns give the share of empty traces (\%).}
\label{tab:change}
\end{table*}

\subsection{Cross-Stage Outcomes and Repair}
\label{app:repair}

The complete cross-stage EM values of Qwen3-4B and Qwen3-14B appear in Table~\ref{tab:full_q3_4b_14b}, and Tables~\ref{tab:contrasts_hotpotqa} and~\ref{tab:contrasts_musique} give paired differences with bootstrap intervals and McNemar tests. The cross-stage rows use the question sets described in Appendix~\ref{app:trace_audit}. Tables~\ref{tab:repair_full} and~\ref{tab:repair_distr} summarize repair separately for the failures of the full-context and the distractor-written baselines.

Because the continuation and separate-call runs of gpt-oss-20b answer from the same gold-written and distractor-written traces, their repair rates can also be compared on the questions that fail under Distr.$\rightarrow$Distr.\ in both runs. This paired comparison confirms that the configuration matters only while the distractor-written trace remains in the input: the separate call repairs far more failures when gold evidence is supplied beside that trace, but the two configurations repair at statistically indistinguishable rates once the trace is rewritten. On the 1,578 such HotpotQA questions, answer-time evidence repairs 4.1\% under continuation and 32.4\% under a separate call (difference $+28.3$ points, 95\% paired bootstrap interval [$+26.0$, $+30.6$]), and rewriting repairs 49.8\% and 49.7\% ($-0.1$ [$-1.3$, $+1.0$]). On the 1,858 such MuSiQue questions, the answer-time rates are 1.6\% and 16.9\% ($+15.3$ [$+13.7$, $+17.0$]), and the rewrite rates are 43.3\% and 42.6\% ($-0.7$ [$-1.6$, $+0.2$]).

\begin{table*}[t]
\centering
\footnotesize
\setlength{\tabcolsep}{2pt}
\begin{tabular}{lccccc}
\toprule
& Answer time & Reasoning time & \multicolumn{3}{c}{Cross-stage} \\
\cmidrule(lr){2-2}\cmidrule(lr){3-3}\cmidrule(lr){4-6}
Model
& \begin{tabular}[b]{@{}c@{}}No trace\\$-$ full\end{tabular}
& \begin{tabular}[b]{@{}c@{}}Gold$\rightarrow$Gold\\$-$ full\end{tabular}
& $n$
& \begin{tabular}[b]{@{}c@{}}Distr.$\rightarrow$Gold\\$-$ Distr.$\rightarrow$Distr.\end{tabular}
& \begin{tabular}[b]{@{}c@{}}Gold$\rightarrow$Distr.\\$-$ Gold$\rightarrow$Gold\end{tabular} \\
\midrule
Qwen3-4B & $-6.5\,[-8.4,-4.6]^{\dag}$ & $+5.9\,[+4.4,+7.5]^{\dag}$ & 1,952 & $+7.1\,[+5.9,+8.4]^{\dag}$ & $-6.0\,[-7.5,-4.7]^{\dag}$ \\
Qwen3-8B & $-4.4\,[-6.2,-2.5]^{\dag}$ & $+4.7\,[+3.0,+6.5]^{\dag}$ & 1,975 & $+19.0\,[+17.2,+20.9]^{\dag}$ & $-18.6\,[-20.6,-16.8]^{\dag}$ \\
Qwen3-14B & $-9.6\,[-11.3,-7.9]^{\dag}$ & $+2.9\,[+1.5,+4.2]^{\dag}$ & 1,999 & $+2.2\,[+1.4,+3.1]^{\dag}$ & $-3.5\,[-4.6,-2.4]^{\dag}$ \\
Qwen3-32B & $-5.8\,[-7.4,-4.2]^{\dag}$ & $+2.8\,[+1.4,+4.2]^{\dag}$ & 764 & $+4.7\,[+2.9,+6.8]^{\dag}$ & $-5.6\,[-7.5,-3.7]^{\dag}$ \\
Qwen3.5-4B & $-7.5\,[-9.1,-6.0]^{\dag}$ & $+2.9\,[+1.4,+4.3]^{\dag}$ & 274 & $+1.5\,[-0.7,+4.0]$ & $-2.2\,[-4.4,-0.4]$ \\
Gemma4-E4B & $-9.1\,[-10.9,-7.4]^{\dag}$ & $+3.5\,[+2.0,+5.0]^{\dag}$ & 1,588 & $+45.0\,[+42.6,+47.5]^{\dag}$ & $-10.3\,[-12.0,-8.6]^{\dag}$ \\
Gemma4-E4B (Cont.) & $-10.7\,[-12.7,-8.8]^{\dag}$ & $+5.8\,[+4.2,+7.6]^{\dag}$ & 1,588 & $+0.3\,[-0.2,+0.8]$ & $-1.0\,[-1.9,-0.1]^{\dag}$ \\
gpt-oss-20b (Cont.) & $-7.7\,[-9.4,-6.0]^{\dag}$ & $+4.1\,[+2.6,+5.6]^{\dag}$ & 2,000 & $+3.4\,[+2.5,+4.4]^{\dag}$ & $-5.1\,[-6.3,-4.0]^{\dag}$ \\
gpt-oss-20b (Sep.) & $-10.8\,[-12.4,-9.1]^{\dag}$ & $+3.1\,[+1.6,+4.6]^{\dag}$ & 2,000 & $+25.2\,[+23.3,+27.2]^{\dag}$ & $-7.4\,[-8.8,-6.0]^{\dag}$ \\
\bottomrule
\end{tabular}
\vspace{-0.05in}
\caption{Paired differences in EM (points) with 95\% bootstrap intervals on HotpotQA. $^{\dag}$ marks an exact McNemar test with $p<0.05$. The first two contrasts use 2,000 questions, except for Gemma4-E4B (Cont.), which uses only questions whose traces are non-empty (1,956 and 1,582). The column $n$ gives the cross-stage sample size.}
\label{tab:contrasts_hotpotqa}
\end{table*}

\begin{table*}[t]
\centering
\footnotesize
\setlength{\tabcolsep}{2pt}
\begin{tabular}{lccccc}
\toprule
& Answer time & Reasoning time & \multicolumn{3}{c}{Cross-stage} \\
\cmidrule(lr){2-2}\cmidrule(lr){3-3}\cmidrule(lr){4-6}
Model
& \begin{tabular}[b]{@{}c@{}}No trace\\$-$ full\end{tabular}
& \begin{tabular}[b]{@{}c@{}}Gold$\rightarrow$Gold\\$-$ full\end{tabular}
& $n$
& \begin{tabular}[b]{@{}c@{}}Distr.$\rightarrow$Gold\\$-$ Distr.$\rightarrow$Distr.\end{tabular}
& \begin{tabular}[b]{@{}c@{}}Gold$\rightarrow$Distr.\\$-$ Gold$\rightarrow$Gold\end{tabular} \\
\midrule
Qwen3-4B & $-11.6\,[-13.5,-9.6]^{\dag}$ & $+14.9\,[+13.0,+17.0]^{\dag}$ & 1,797 & $+3.6\,[+2.6,+4.5]^{\dag}$ & $-5.9\,[-7.3,-4.5]^{\dag}$ \\
Qwen3-8B & $-12.8\,[-14.9,-11.0]^{\dag}$ & $+12.0\,[+10.0,+14.1]^{\dag}$ & 1,891 & $+12.2\,[+10.8,+13.7]^{\dag}$ & $-7.3\,[-8.8,-5.8]^{\dag}$ \\
Qwen3-14B & $-10.4\,[-12.5,-8.6]^{\dag}$ & $+12.4\,[+10.5,+14.4]^{\dag}$ & 1,991 & $+2.2\,[+1.5,+3.0]^{\dag}$ & $-1.9\,[-2.9,-1.0]^{\dag}$ \\
Qwen3-32B & $-9.8\,[-11.3,-8.2]^{\dag}$ & $+10.9\,[+9.1,+12.8]^{\dag}$ & 422 & $+3.6\,[+1.7,+5.5]^{\dag}$ & $-4.3\,[-6.6,-1.9]^{\dag}$ \\
Qwen3.5-4B & $-11.0\,[-12.7,-9.4]^{\dag}$ & $+10.6\,[+8.5,+12.7]^{\dag}$ & 61 & $+1.6\,[+0.0,+4.9]$ & $+1.6\,[-3.3,+8.2]$ \\
Gemma4-E4B & $-15.5\,[-17.5,-13.5]^{\dag}$ & $+12.3\,[+10.4,+14.2]^{\dag}$ & 1,889 & $+22.2\,[+20.4,+24.1]^{\dag}$ & $-8.5\,[-9.8,-7.1]^{\dag}$ \\
Gemma4-E4B (Cont.) & $-16.3\,[-18.3,-14.4]^{\dag}$ & $+11.3\,[+9.4,+13.3]^{\dag}$ & 1,889 & $+0.5\,[+0.2,+0.8]^{\dag}$ & $-1.3\,[-2.2,-0.5]^{\dag}$ \\
gpt-oss-20b (Cont.) & $-17.0\,[-18.9,-15.1]^{\dag}$ & $+8.4\,[+6.6,+10.4]^{\dag}$ & 2,000 & $+1.6\,[+1.0,+2.2]^{\dag}$ & $-3.2\,[-4.1,-2.3]^{\dag}$ \\
gpt-oss-20b (Sep.) & $-18.4\,[-20.2,-16.4]^{\dag}$ & $+6.8\,[+5.0,+8.7]^{\dag}$ & 2,000 & $+15.7\,[+14.2,+17.5]^{\dag}$ & $-3.3\,[-4.4,-2.2]^{\dag}$ \\
\bottomrule
\end{tabular}
\vspace{-0.05in}
\caption{Paired differences in EM (points) with 95\% bootstrap intervals on MuSiQue. $^{\dag}$ marks an exact McNemar test with $p<0.05$. The first two contrasts use 2,000 questions, except for Gemma4-E4B (Cont.), which uses only questions whose traces are non-empty (1,994 and 1,889). The column $n$ gives the cross-stage sample size.}
\label{tab:contrasts_musique}
\end{table*}

\begin{table*}[t]
\begin{minipage}[t]{\columnwidth}
\vspace{0pt}
\centering
\small
\setlength{\tabcolsep}{2.5pt}
\begin{tabular}{lccc}
\toprule
Model & $n$ & Gold at answer time & Rewrite \\
\midrule
\multicolumn{4}{l}{\textit{HotpotQA}} \\
Qwen3-4B (Cont.) & 1,039 & 8.9 & 18.3 \\
Qwen3-8B (Cont.) & 979 & 10.7 & 21.1 \\
Qwen3-14B (Cont.) & 936 & 4.6 & 13.7 \\
Qwen3-32B (Cont.) & 795 & 5.0 & 12.2 \\
gpt-oss-20b (Cont.) & 917 & 5.1 & 17.3 \\
gpt-oss-20b (Sep.) & 885 & 7.9 & 17.1 \\
Qwen3.5-4B (Sep.) & 404 & 1.7 & 10.4 \\
Gemma4-E4B (Sep.) & 662 & 3.8 & 20.8 \\
\midrule
\multicolumn{4}{l}{\textit{MuSiQue}} \\
Qwen3-4B (Cont.) & 1,407 & 7.3 & 27.4 \\
Qwen3-8B (Cont.) & 1,262 & 9.0 & 26.9 \\
Qwen3-14B (Cont.) & 1,282 & 3.7 & 25.8 \\
Qwen3-32B (Cont.) & 548 & 6.0 & 16.1 \\
gpt-oss-20b (Cont.) & 1,253 & 3.2 & 21.5 \\
gpt-oss-20b (Sep.) & 1,235 & 3.1 & 20.2 \\
Qwen3.5-4B (Sep.) & 168 & 0.6 & 23.8 \\
Gemma4-E4B (Sep.) & 1,205 & 6.4 & 26.3 \\
\bottomrule
\end{tabular}
\caption{Repair rates (\%) among failures of standard inference (Full$\rightarrow$Full). Gold at answer time removes distractors (Full$\rightarrow$Gold), and Rewrite generates a new trace using gold evidence (Gold$\rightarrow$Gold). $n$ is the number of failed questions for each model and configuration (question sets in Appendix~\ref{app:trace_audit}).}
\label{tab:repair_full}
\end{minipage}\hfill
\begin{minipage}[t]{\columnwidth}
\vspace{0pt}
\centering
\small
\setlength{\tabcolsep}{2.5pt}
\begin{tabular}{lccc}
\toprule
Model & $n$ & Gold at answer time & Rewrite \\
\midrule
\multicolumn{4}{l}{\textit{HotpotQA}} \\
Qwen3-4B (Cont.) & 1,777 & 8.6 & 50.5 \\
Qwen3-8B (Cont.) & 1,740 & 22.9 & 51.6 \\
Qwen3-14B (Cont.) & 1,789 & 3.5 & 52.2 \\
Qwen3-32B (Cont.) & 583 & 8.2 & 45.6 \\
gpt-oss-20b (Cont.) & 1,602 & 5.1 & 50.0 \\
gpt-oss-20b (Sep.) & 1,595 & 32.7 & 49.8 \\
Qwen3.5-4B (Sep.) & 131 & 6.1 & 55.7 \\
Gemma4-E4B (Sep.) & 1,481 & 48.9 & 61.0 \\
\midrule
\multicolumn{4}{l}{\textit{MuSiQue}} \\
Qwen3-4B (Cont.) & 1,757 & 3.9 & 44.6 \\
Qwen3-8B (Cont.) & 1,827 & 12.8 & 45.9 \\
Qwen3-14B (Cont.) & 1,938 & 2.5 & 47.4 \\
Qwen3-32B (Cont.) & 389 & 4.1 & 47.8 \\
gpt-oss-20b (Cont.) & 1,864 & 1.8 & 43.4 \\
gpt-oss-20b (Sep.) & 1,872 & 17.2 & 42.6 \\
Qwen3.5-4B (Sep.) & 52 & 1.9 & 57.7 \\
Gemma4-E4B (Sep.) & 1,868 & 22.5 & 48.7 \\
\bottomrule
\end{tabular}
\caption{Repair rates (\%) among failures of the distractor-written run (Distr.$\rightarrow$Distr.). Gold at answer time keeps the distractor-written trace and supplies gold evidence (Distr.$\rightarrow$Gold), and Rewrite answers from the gold-written trace (Gold$\rightarrow$Gold). $n$ is the number of Distr.$\rightarrow$Distr.\ failures in the run of each row (question sets and runs in Appendix~\ref{app:trace_audit}).}
\label{tab:repair_distr}
\end{minipage}
\end{table*}

\subsection{Complete Per-Model Metrics}

Tables~\ref{tab:full_q3_4b_14b} through~\ref{tab:full_q35_g4} report EM, F1, and Contains for the main answer-time and reasoning-time conditions of the Qwen and Gemma models on both datasets, and Table~\ref{tab:full_q3_4b_14b} adds the cross-stage conditions of Qwen3-4B and Qwen3-14B. Appendix~\ref{app:deepseek} reports DeepSeek-V4-Pro, and the gpt-oss-20b conditions are reported only in the contrast and repair tables.

\begin{table*}[t]
\centering
\footnotesize
\setlength{\tabcolsep}{3pt}
\begin{tabular}{lcccccccccccc}
\toprule
 & \multicolumn{6}{c}{Qwen3-4B} & \multicolumn{6}{c}{Qwen3-14B} \\
\cmidrule(lr){2-7}\cmidrule(lr){8-13}
 & \multicolumn{3}{c}{HotpotQA} & \multicolumn{3}{c}{MuSiQue} & \multicolumn{3}{c}{HotpotQA} & \multicolumn{3}{c}{MuSiQue} \\
\cmidrule(lr){2-4}\cmidrule(lr){5-7}\cmidrule(lr){8-10}\cmidrule(lr){11-13}
Condition & EM & F1 & Contains & EM & F1 & Contains & EM & F1 & Contains & EM & F1 & Contains \\
\midrule
Trace + gold + distr. & 48.0 & 66.1 & 64.5 & 29.6 & 43.4 & 39.0 & 53.2 & 69.6 & 65.9 & 35.9 & 49.3 & 44.9 \\
Gold + distr., no trace & 41.5 & 55.8 & 50.9 & 18.1 & 29.6 & 22.8 & 43.6 & 59.0 & 55.5 & 25.5 & 38.1 & 31.5 \\
Half trace + gold + distr. & 46.8 & 63.7 & 61.3 & 27.7 & 40.7 & 34.8 & 50.5 & 66.8 & 63.5 & 34.1 & 46.9 & 41.6 \\
Trace only & 50.7 & 67.8 & 63.3 & 30.3 & 43.8 & 37.3 & 50.6 & 67.7 & 64.2 & 33.5 & 47.6 & 44.2 \\
Trace + distr. & 44.7 & 62.8 & 62.3 & 27.0 & 40.8 & 36.6 & 49.5 & 65.6 & 62.4 & 33.1 & 46.7 & 42.4 \\
Trace + gold & 50.2 & 67.5 & 64.7 & 33.2 & 47.3 & 41.7 & 53.6 & 70.0 & 66.6 & 37.2 & 50.6 & 46.8 \\
Nothing & 13.3 & 20.4 & 14.1 & 1.8 & 9.3 & 1.9 & 16.1 & 24.2 & 17.9 & 3.5 & 11.1 & 3.5 \\
Question-only trace and answer & 11.2 & 17.3 & 18.1 & 3.0 & 7.7 & 3.7 & 14.0 & 20.9 & 25.3 & 3.9 & 8.6 & 6.5 \\
\midrule
Gold$\rightarrow$Gold & 54.0 & 71.8 & 67.8 & 44.6 & 60.2 & 54.3 & 56.0 & 72.4 & 69.2 & 48.3 & 63.4 & 58.7 \\
Distr.$\rightarrow$Distr. & 8.6 & 16.0 & 15.5 & 1.9 & 7.5 & 3.4 & 10.4 & 16.2 & 14.5 & 2.7 & 7.9 & 3.8 \\
Gold$\rightarrow$Full & 51.2 & 69.8 & 67.8 & 41.3 & 56.8 & 52.1 & 55.3 & 71.8 & 68.9 & 47.5 & 62.6 & 57.5 \\
Gold$\rightarrow$Distr. & 48.2 & 67.1 & 65.7 & 39.0 & 55.0 & 50.9 & 52.5 & 69.2 & 65.8 & 46.5 & 61.3 & 56.6 \\
Distr.$\rightarrow$Gold & 17.0 & 25.7 & 25.7 & 8.2 & 15.5 & 11.9 & 12.7 & 18.2 & 17.3 & 5.0 & 10.1 & 7.1 \\
Distr.$\rightarrow$Full & 16.1 & 25.4 & 25.6 & 5.2 & 11.8 & 8.0 & 13.9 & 20.0 & 18.4 & 4.5 & 9.6 & 6.1 \\
\bottomrule
\end{tabular}
\vspace{-0.05in}
\caption{EM, F1, and Contains (\%) for Qwen3-4B and Qwen3-14B on 2,000 questions per dataset. The conditions follow Tables~\ref{tab:post} and~\ref{tab:pre}, and the last four rows are the cross-stage conditions of Table~\ref{tab:conditions}.}
\label{tab:full_q3_4b_14b}
\end{table*}

\begin{table*}[t]
\centering
\footnotesize
\setlength{\tabcolsep}{3pt}
\begin{tabular}{lcccccccccccc}
\toprule
 & \multicolumn{6}{c}{Qwen3-8B} & \multicolumn{6}{c}{Qwen3-32B} \\
\cmidrule(lr){2-7}\cmidrule(lr){8-13}
 & \multicolumn{3}{c}{HotpotQA} & \multicolumn{3}{c}{MuSiQue} & \multicolumn{3}{c}{HotpotQA} & \multicolumn{3}{c}{MuSiQue} \\
\cmidrule(lr){2-4}\cmidrule(lr){5-7}\cmidrule(lr){8-10}\cmidrule(lr){11-13}
Condition & EM & F1 & Contains & EM & F1 & Contains & EM & F1 & Contains & EM & F1 & Contains \\
\midrule
Trace + gold + distr. & 50.9 & 68.0 & 63.7 & 35.0 & 48.3 & 40.9 & 53.2 & 70.7 & 66.5 & 38.6 & 53.3 & 47.7 \\
Gold + distr., no trace & 46.6 & 60.7 & 53.1 & 22.2 & 35.3 & 25.0 & 47.4 & 63.3 & 59.8 & 28.9 & 43.0 & 35.7 \\
Half trace + gold + distr. & 50.3 & 66.8 & 61.3 & 32.4 & 46.2 & 38.3 & 50.9 & 68.6 & 65.0 & 37.1 & 51.7 & 46.2 \\
Trace only & 37.6 & 51.9 & 46.9 & 29.3 & 41.9 & 34.3 & 49.8 & 68.0 & 63.4 & 29.5 & 44.4 & 39.8 \\
Trace + distr. & 35.6 & 50.1 & 45.2 & 27.9 & 41.0 & 33.4 & 47.7 & 65.1 & 61.9 & 28.9 & 42.7 & 38.9 \\
Trace + gold & 52.8 & 70.2 & 66.4 & 39.5 & 53.2 & 45.8 & 53.2 & 71.1 & 67.9 & 46.6 & 63.8 & 58.1 \\
Nothing & 14.8 & 22.9 & 15.7 & 2.6 & 11.0 & 3.1 & 17.7 & 26.7 & 19.4 & 4.0 & 13.5 & 4.5 \\
Question-only trace and answer & 14.2 & 21.4 & 19.7 & 2.9 & 8.7 & 4.8 & 13.2 & 20.5 & 27.7 & 3.7 & 8.7 & 7.3 \\
\midrule
Gold$\rightarrow$Gold & 55.7 & 72.6 & 67.7 & 47.0 & 62.3 & 54.9 & 56.0 & 73.4 & 70.2 & 49.5 & 67.2 & 61.3 \\
Distr.$\rightarrow$Distr. & 11.8 & 19.6 & 16.5 & 3.3 & 10.0 & 4.8 & 14.6 & 23.0 & 19.4 & 3.2 & 11.2 & 4.6 \\
\bottomrule
\end{tabular}
\vspace{-0.05in}
\caption{EM, F1, and Contains (\%) for Qwen3-8B and Qwen3-32B on 2,000 questions per dataset. The conditions follow Tables~\ref{tab:post} and~\ref{tab:pre}.}
\label{tab:full_q3_8b_32b}
\end{table*}

\begin{table*}[t]
\centering
\footnotesize
\setlength{\tabcolsep}{3pt}
\begin{tabular}{lcccccccccccc}
\toprule
 & \multicolumn{6}{c}{Qwen3.5-4B} & \multicolumn{6}{c}{Gemma4-E4B} \\
\cmidrule(lr){2-7}\cmidrule(lr){8-13}
 & \multicolumn{3}{c}{HotpotQA} & \multicolumn{3}{c}{MuSiQue} & \multicolumn{3}{c}{HotpotQA} & \multicolumn{3}{c}{MuSiQue} \\
\cmidrule(lr){2-4}\cmidrule(lr){5-7}\cmidrule(lr){8-10}\cmidrule(lr){11-13}
Condition & EM & F1 & Contains & EM & F1 & Contains & EM & F1 & Contains & EM & F1 & Contains \\
\midrule
Trace + gold + distr. & 61.8 & 74.8 & 65.2 & 38.5 & 47.4 & 37.7 & 58.6 & 75.8 & 72.4 & 37.2 & 50.4 & 45.3 \\
Gold + distr., no trace & 54.3 & 67.6 & 56.6 & 27.5 & 37.5 & 25.4 & 49.5 & 64.4 & 60.5 & 21.8 & 32.3 & 26.3 \\
Half trace + gold + distr. & 62.5 & 75.8 & 65.8 & 39.6 & 48.9 & 38.8 & 58.3 & 74.5 & 70.8 & 36.4 & 49.7 & 44.0 \\
Trace only & 56.2 & 67.5 & 59.9 & 25.4 & 29.9 & 28.8 & 57.0 & 72.2 & 66.9 & 31.7 & 41.0 & 35.4 \\
Trace + distr. & 56.8 & 67.8 & 59.5 & 27.6 & 32.4 & 28.2 & 52.9 & 68.1 & 63.7 & 33.6 & 45.2 & 39.3 \\
Trace + gold & 63.8 & 76.8 & 66.8 & 44.8 & 54.6 & 45.4 & 58.6 & 76.0 & 72.7 & 40.6 & 53.9 & 49.3 \\
Nothing & 17.9 & 25.2 & 18.4 & 3.6 & 10.0 & 3.8 & 16.1 & 23.1 & 17.3 & 2.6 & 7.3 & 2.7 \\
Question-only trace and answer & 14.5 & 18.2 & 14.8 & 2.6 & 4.4 & 2.6 & 13.9 & 19.8 & 15.4 & 2.7 & 6.0 & 3.0 \\
\midrule
Gold$\rightarrow$Gold & 64.7 & 78.2 & 68.6 & 49.1 & 58.4 & 51.5 & 62.1 & 78.4 & 73.9 & 49.5 & 64.5 & 60.6 \\
Distr.$\rightarrow$Distr. & 15.0 & 19.8 & 16.2 & 2.6 & 6.2 & 2.7 & 7.7 & 11.8 & 9.8 & 1.2 & 4.3 & 1.8 \\
\bottomrule
\end{tabular}
\vspace{-0.05in}
\caption{EM, F1, and Contains (\%) for Qwen3.5-4B and Gemma4-E4B on 2,000 questions per dataset. The conditions follow Tables~\ref{tab:post} and~\ref{tab:pre}.}
\label{tab:full_q35_g4}
\end{table*}

\clearpage\fullcolumn
\section{Supplementary Experiments}
\label{app:audits}

All accuracy and outcome values in this appendix are percentages.

\subsection{Information Carried by the Trace}

Table~\ref{tab:controls} separates the final line of the model's full-context trace from the preceding steps and compares both inputs with alternative texts. Final line only and Trace minus final line partition the trace. The extractive summary lists gold supporting sentences, using the sentences that contain each hop answer within MuSiQue supporting paragraphs, and is cut to the token length of the model's own trace. This condition therefore relies on oracle evidence selection. The unrelated trace is a length-matched trace that comes from another question and does not contain the gold answer. The two no-trace controls use the placeholder from the main runs or an empty trace field. The own-trace and placeholder rows were generated together with the other controls and differ from Table~\ref{tab:post} by up to 0.3 points.

Section~\ref{sec:follow} discusses the final-line rows: the final line alone stays within 2.0 EM points of the own trace, while removing it costs 2.9 to 8.5 points. The extractive summary differs from the full trace by at most 1.5 EM points on HotpotQA and exceeds it by 2.2 to 2.7 points on MuSiQue, so a short text that states the needed facts can take the place of the trace.

\begin{table}[t]
\centering
\footnotesize
\setlength{\tabcolsep}{3pt}
\begin{tabular}{lcccc}
\toprule
& \multicolumn{2}{c}{HotpotQA} & \multicolumn{2}{c}{MuSiQue} \\
\cmidrule(lr){2-3}\cmidrule(lr){4-5}
Trace given & 4B & 14B & 4B & 14B \\
\midrule
Own trace & 48.0 & 53.1 & 29.9 & 35.9 \\
Final line only & 46.0$^{\dag}$ & 52.2 & 30.0 & 34.7$^{\dag}$ \\
Trace minus final line & 42.1$^{\dag}$ & 50.2$^{\dag}$ & 21.4$^{\dag}$ & 32.2$^{\dag}$ \\
Extractive summary & 47.5 & 51.6 & 32.6$^{\dag}$ & 38.1$^{\dag}$ \\
Unrelated trace & 7.2$^{\dag}$ & 8.7$^{\dag}$ & 1.1$^{\dag}$ & 1.2$^{\dag}$ \\
No trace (placeholder) & 41.8$^{\dag}$ & 43.7$^{\dag}$ & 18.2$^{\dag}$ & 25.6$^{\dag}$ \\
No trace (empty) & 37.9$^{\dag}$ & 43.3$^{\dag}$ & 16.1$^{\dag}$ & 24.5$^{\dag}$ \\
\bottomrule
\end{tabular}
\caption{Trace-content controls for Qwen3-4B and Qwen3-14B with full answer-time evidence (2,000 questions, EM in \%). $^{\dag}$ marks a significant difference from the model's own trace (McNemar, $p<0.05$).}
\label{tab:controls}
\end{table}

\subsection{Sensitivity to Trace Corruption}

Table~\ref{tab:liveness} edits the model's full-context trace while the answer stage continues to receive the full evidence. Answer swap replaces the gold-answer string in the trace with the title of a distractor paragraph and applies to traces containing that string. Step deletion removes the step most similar to a gold supporting sentence. To separate the effect of a content edit from that of mere rewording, the Sel.\ column subtracts the share of answers broken by a Qwen3-1.7B paraphrase of the trace (prompt in Appendix~\ref{app:prompts}). The paraphrase control uses only paraphrases that pass an entity check, which requires the paraphrase to keep the entities and numbers of the trace. Most Qwen3-32B paraphrases fail this check, so the Qwen3-32B paraphrase control rests on only 118 HotpotQA and 44 MuSiQue questions.

\begin{table}[t]
\centering
\footnotesize
\setlength{\tabcolsep}{2.4pt}
\begin{tabular}{lccccccc}
\toprule
& & & \multicolumn{3}{c}{Answer swap} & \multicolumn{2}{c}{Step deletion} \\
\cmidrule(lr){4-6}\cmidrule(lr){7-8}
Qwen3 & $n$ & $n_{\textrm{para}}$ & Broken & Sel. & Adopt & Broken & Sel. \\
\midrule
\multicolumn{8}{l}{\textit{HotpotQA}} \\
4B & 961 & 745 & 51.1 & +48.0 & 29.6 & 11.1 & +8.0 \\
8B & 1,012 & 859 & 57.7 & +56.9 & 38.9 & 5.4 & +4.6 \\
14B & 1,064 & 796 & 50.7 & +49.8 & 30.8 & 5.5 & +4.6 \\
32B & 1,014 & 118 & 39.9 & +38.2 & 27.5 & 5.4 & +3.7 \\
\midrule
\multicolumn{8}{l}{\textit{MuSiQue}} \\
4B & 590 & 436 & 63.5 & +59.0 & 35.6 & 16.3 & +11.7 \\
8B & 694 & 468 & 63.6 & +62.8 & 41.8 & 8.7 & +7.8 \\
14B & 713 & 434 & 60.8 & +59.6 & 38.5 & 8.0 & +6.8 \\
32B & 608 & 44 & 46.9 & +44.6 & 30.5 & 7.7 & +5.4 \\
\bottomrule
\end{tabular}
\caption{Trace corruption for the four Qwen3 sizes, on the questions whose full-context trace is non-empty and whose standard-inference answer is correct ($n$). Broken is the share of answers that become incorrect (\%). Sel.\ (selectivity) is Broken minus the share broken by a paraphrase of the trace and is computed on the $n_{\textrm{para}}$ questions whose paraphrase passes the entity check. Adopt is the share of swapped cases whose answer is the inserted title. For every model and dataset, swapping the answer string breaks a far larger share of answers than step deletion.}
\label{tab:liveness}
\end{table}

\subsection{Text Deletion and Cache Masking}

Text deletion changes the length and positions of the remaining input, while cache masking preserves positions and removes selected segments through the key-value cache of a single pass over the prompt and trace. Table~\ref{tab:kv_replication} reports both implementations for Qwen3-4B and Qwen3-14B and thus indicates whether their answer-time ablation results depend on the length and position changes that deletion introduces. Without masking, the single pass gives the standard answer for 90.2\% to 97.2\% of questions. The main patterns hold under both implementations: removing the trace lowers EM, the trace alone gives EM similar to that of the full input, and trace plus gold evidence scores higher than trace plus distractors.

\begin{table}[t]
\centering
\footnotesize
\setlength{\tabcolsep}{3pt}
\begin{tabular}{lcccc}
\toprule
& \multicolumn{2}{c}{Text deletion} & \multicolumn{2}{c}{Cache masking} \\
\cmidrule(lr){2-3}\cmidrule(lr){4-5}
Answer-stage input & 4B & 14B & 4B & 14B \\
\midrule
\multicolumn{5}{l}{\textit{HotpotQA}} \\
Trace + gold + distr. & 48.0 & 53.2 & 48.4 & 53.3 \\
Gold + distr., no trace & 41.5 & 43.6 & 40.0 & 43.4 \\
Trace only & 50.7 & 50.6 & 49.2 & 52.0 \\
Trace + distr. & 44.7 & 49.5 & 47.6 & 52.1 \\
Trace + gold & 50.2 & 53.6 & 49.6 & 53.5 \\
\midrule
\multicolumn{5}{l}{\textit{MuSiQue}} \\
Trace + gold + distr. & 29.6 & 35.9 & 29.9 & 35.7 \\
Gold + distr., no trace & 18.1 & 25.5 & 16.8 & 24.7 \\
Trace only & 30.3 & 33.5 & 29.6 & 35.3 \\
Trace + distr. & 27.0 & 33.1 & 28.9 & 34.7 \\
Trace + gold & 33.2 & 37.2 & 30.8 & 36.4 \\
\bottomrule
\end{tabular}
\caption{Answer-time ablations of Qwen3-4B and Qwen3-14B with two implementations: deleting segments from the prompt text, as in the main runs, and masking the same segments in the key-value cache of a single pass over prompt and trace (positions unchanged). EM (\%), 2,000 questions.}
\label{tab:kv_replication}
\end{table}

\subsection{DeepSeek-V4-Pro Through Its API}
\label{app:deepseek}

DeepSeek-V4-Pro is larger than the other models, and we query it through its API. All queries were made in October 2026 (system fingerprint \texttt{a307abda}), and Appendices~\ref{app:setup} and~\ref{app:prompts} give the limits and prompts. All of its runs use HotpotQA and answer with a separate call, so it does not enter the comparison between continuation and separate calls. The answer-time ablations use the first 1,205 validation questions. The conflict experiment uses the first 200 verified evidence pairs, and trace filtering retains 149 of them. Of the 51 removed pairs, 48 are removed only because the trace written from $E_1$ also names $a_0$. None of its traces is empty, and 87.0\% contain the gold answer. Table~\ref{tab:deepseek} has no Half trace row, since 90.1\% of these traces are a single line that Half trace would remove entirely.

Under a separate call, the answers of DeepSeek-V4-Pro rely substantially on the trace (Table~\ref{tab:deepseek}): removing the trace lowers EM from 62.5 to 57.4 and changes 23.9\% of the answers, while removing all evidence changes only 6.9\%. In the crossed conflicts, the separate call follows the evidence in 58.4\% of the answers with gold evidence only and in 53.4\% with the original distractors added, a decline in the same direction as in Section~\ref{sec:noise_boundary}. When the stale trace states the counterfactual $a_1$, the model follows the trace in 16.1\% of the answers and the evidence in 77.9\%. When the stale trace states the original answer $a_0$, it follows the trace in 54.4\% and the evidence in 38.9\%. This difference between the directions is larger than for any model in Table~\ref{tab:conflict_dir}, and knowledge of $a_0$ can raise trace-following when the trace states $a_0$ (Appendix~\ref{app:planned}).

\begin{table}[t]
\centering
\footnotesize
\setlength{\tabcolsep}{2.3pt}
\begin{tabular}{lcccc}
\toprule
\multicolumn{5}{l}{\textit{Answer-time ablations, first 1,205 questions}} \\
Answer input & EM & F1 & Contains & Changed \\
\midrule
Trace + gold + distr. & 62.5 & 80.0 & 73.4 & -- \\
Gold + distr., no trace & 57.4 & 74.2 & 68.0 & 23.9 \\
Trace only & 63.7 & 80.6 & 72.9 & 6.9 \\
Trace + distr. & 62.4 & 79.8 & 72.9 & -- \\
Trace + gold & 63.1 & 80.3 & 73.6 & 6.2 \\
Nothing & 31.0 & 44.2 & 36.9 & -- \\
\midrule
\multicolumn{5}{l}{\textit{Crossed conflicts, 149 retained pairs}} \\
Evidence & Matched & Trace & Evid. & Other \\
\midrule
Gold only & 92.6 & 35.2 & 58.4 & 6.4 \\
Gold + distr. & 93.3 & 38.9 & 53.4 & 7.7 \\
\bottomrule
\end{tabular}
\caption{DeepSeek-V4-Pro on HotpotQA with a separate answer call. All values are percentages. Changed is the share of answers that differ from the full-input answer when the trace, all evidence, or only the distractors are removed. The conflict columns follow Table~\ref{tab:conflict}.}
\label{tab:deepseek}
\end{table}

\section{Prompts}
\label{app:prompts}

This appendix reproduces the prompts verbatim. Fields in braces are instantiated for each question, and Section~\ref{sec:inference} describes the continuation and separate-call structures.

\paragraph{Qwen3 (raw prompt, no chat template).}
The reasoning call completes the prompt below, and the trace is the text before \texttt{</think>}. Continuation appends the (possibly ablated) trace and \texttt{\textbackslash n</think>\textbackslash n<answer>} to the same prompt and reads the answer up to \texttt{</answer>}.
\begin{promptbox}{Qwen3 reasoning call}
You are a precise multi-hop QA assistant.
Read the EVIDENCE and answer the QUESTION.
You MUST reason in multiple concise steps before final answer.
Output format must be EXACTLY:
<think>
1) step one
2) step two
...
</think>
<answer>short final answer phrase</answer>
No text outside these tags.

QUESTION:
{question}

EVIDENCE:
{evidence}

OUTPUT:
<think>
\end{promptbox}
The question-only call uses the same format without evidence, with the text below.
\begin{promptbox}{Qwen3 question-only call}
No external evidence is provided. Answer the QUESTION from your own knowledge. Reason 4-6 steps inside <think></think> tags, then give the final answer inside <answer></answer> tags.
\end{promptbox}
\emph{Verify} inserts the text below after ``Read the EVIDENCE and answer the QUESTION.''
\begin{promptbox}{Verify}
The reasoning may be outdated. Before giving the final answer, check each step against the EVIDENCE; if the reasoning and the EVIDENCE disagree, answer from the EVIDENCE.
\end{promptbox}
\emph{Re-reason} runs a new reasoning call whose QUESTION field is the question followed by the notes below, and the answer of this call is the output.
\begin{promptbox}{Re-reason, QUESTION field}
{question}

PREVIOUS NOTES (may be outdated or wrong; re-check them against the EVIDENCE):
{trace}
\end{promptbox}

\paragraph{Gemma4-E4B, gpt-oss-20b, and the separate call of Qwen3-14B (chat templates).}
The reasoning call is one user turn with thinking on (Gemma4-E4B also has the system message ``You are a concise QA assistant.'', and gpt-oss-20b uses low reasoning effort and no system message).
\begin{promptbox}{Reasoning call}
Answer the QUESTION with a short factual phrase (whatever the question asks for).
Think briefly.

QUESTION:
{question}

EVIDENCE:
{evidence}
\end{promptbox}
In continuation, the answer stage repeats this prompt and places the trace in the model's own thinking block (for Gemma4-E4B \texttt{<|channel>thought\textbackslash n\allowbreak\{trace\}\allowbreak\textbackslash n<channel|>}, and for gpt-oss-20b the analysis channel, closed by \texttt{<|end|>\allowbreak<|start|>\allowbreak assistant\allowbreak<|channel|>\allowbreak final\allowbreak<|message|>}). The separate call is a new request with thinking off.
\begin{promptbox}{Separate answer call}
Given the REASONING TRACE and EVIDENCE below (even if they are empty),Answer the
QUESTION with a short factual phrase (whatever the question asks for). QUESTION:
{question}

REASONING TRACE:
{trace}

EVIDENCE:
{evidence}
\end{promptbox}
The variants of Appendix~\ref{app:components} change only the block order (EVIDENCE before REASONING TRACE) or the trace header.
\begin{promptbox}{Trace headers of the component variants}
REASONING TRACE (your own earlier reasoning for this question):
REASONING TRACE (written by another model for this question):
\end{promptbox}
Qwen3.5-4B uses the same structure with the instruction and the system message below.
\begin{promptbox}{Qwen3.5-4B instruction}
Answer the QUESTION with a short factual phrase like Yes or No, or just the name or a number (whatever the question asks for)
\end{promptbox}
\begin{promptbox}{Qwen3.5-4B system message}
You are a concise QA assistant. When thinking, do NOT restate the question, analyze constraints, or do a final check. Think directly about the answer in 2-3 short steps, then respond.
\end{promptbox}

\paragraph{DeepSeek-V4-Pro (API).}
The reasoning call (thinking enabled) sends the first text below followed by the QUESTION and EVIDENCE fields, and the trace is the returned reasoning content. The answer call (thinking disabled) sends the second text followed by QUESTION, REASONING TRACE, and EVIDENCE.
\begin{promptbox}{DeepSeek-V4-Pro reasoning call}
Answer the QUESTION based on the EVIDENCE. Output only: <answer>your answer</answer>
\end{promptbox}
\begin{promptbox}{DeepSeek-V4-Pro answer call}
Given the REASONING TRACE and EVIDENCE below, answer the QUESTION with a short factual phrase even if the reasoning trace and evidence are empty. Output only: <answer>your answer</answer>
\end{promptbox}

\paragraph{Counterfactual proposer and verifier.}
The proposer of Appendix~\ref{app:planned} (gemma-3-27b-it) receives the first prompt below as a user turn, where \texttt{\{answer\}} is the original answer $a_0$. The verifier answers each evidence version with the second prompt, and each gold paragraph is rendered as its title followed by its text.
\begin{promptbox}{Proposer (gemma-3-27b-it)}
Question: {question}
Correct answer: {answer}

Name ONE different real-world entity of exactly the same type as the correct answer (another person if it is a person, another city if it is a city, another year if it is a year, another film if it is a film, and so on). It must be a plausible answer to the question but different from the correct answer, and it must not be a part, variant or alias of it. Reply with the entity name only.
\end{promptbox}
\begin{promptbox}{Verifier (gemma-3-27b-it)}
Answer the question using only the passages below, even if they contradict what you know. Reply with the shortest possible answer phrase only.

PASSAGES:
{passages}

QUESTION: {question}
\end{promptbox}

\paragraph{Trace paraphrase.}
The paraphrase control of the trace-corruption experiment (Table~\ref{tab:liveness}) uses Qwen3-1.7B with the prompt below, where \texttt{\{steps\}} holds the steps of the trace, one per line.
\begin{promptbox}{Paraphrase (Qwen3-1.7B)}
Rewrite the following reasoning steps in different words. Keep every name, entity and number EXACTLY unchanged, keep one step per line with the same numbering, and do not add or remove steps.

STEPS:
{steps}

REWRITTEN STEPS:
\end{promptbox}

\end{document}